\documentclass{article}

\PassOptionsToPackage{numbers,compress}{natbib}

\usepackage[main,final]{neurips_2026}

\usepackage[utf8]{inputenc}
\usepackage[T1]{fontenc}

\ifdefined\XeTeXversion
  \usepackage{newunicodechar}
  \newunicodechar{’}{'}
  \newunicodechar{“}{``}
  \newunicodechar{”}{''}
  \newunicodechar{–}{--}
\fi

\usepackage{graphicx}
\usepackage{subcaption}
\usepackage{multirow}
\usepackage{makecell}
\usepackage{tabularx}
\usepackage{caption}
\usepackage{array}
\usepackage{wrapfig}
\usepackage{mathtools}
\usepackage{amsthm}
\usepackage{amsmath}
\usepackage{amssymb}
\usepackage[table]{xcolor}
\usepackage{enumitem}
\usepackage[textsize=tiny]{todonotes}
\usepackage{tcolorbox}
\usepackage{booktabs}
\usepackage{hyperref}
\usepackage{xurl}
\usepackage[capitalize,noabbrev]{cleveref}
\usepackage{url}
\usepackage{algorithm}
\usepackage{algorithmic}

\definecolor{comment_color_2}{RGB}{64,128,128}

\newcommand{\LineComment}[1]{%
  \vspace*{0.5em}\small\textcolor{comment_color_2}{\textit{\# #1}}%
}

\newcommand{\tscell}[1]{\cellcolor{cyan!10}#1}

\theoremstyle{plain}

\theoremstyle{definition}

\theoremstyle{remark}

\newtcolorbox{conclusionbox}{
  colback=gray!5,
  colframe=gray!60,
  boxrule=0.8pt,
  arc=2pt,
  left=6pt,
  right=6pt,
  top=4pt,
  bottom=4pt
}

\title{TriSpec: Ternary Speculative Decoding via Lightweight Proxy Verification}

\author{%
  \parbox{\dimexpr\textwidth-2\tabcolsep\relax}{\centering
    \mbox{\textbf{Haoyun Jiang}$^{1,2}$} \quad
    \mbox{\textbf{Junqi He}$^{1}$} \quad
    \mbox{\textbf{Feng Hong}$^{1}$} \quad
    \mbox{\textbf{Xinlong Yang}$^{2}$} \quad
    \mbox{\textbf{Jianwei Zhang}$^{2}$} \quad
    \mbox{\textbf{Zhengyang Zhuge}$^{2}$} \quad
    \mbox{\textbf{Zheng Li}$^{2}$} \quad
    \mbox{\textbf{Xiaofeng Cao}$^{3}$} \quad
    \mbox{\textbf{Zhiyong Chen}$^{1}$} \quad
    \mbox{\textbf{Bo Han}$^{4}$} \quad
    \mbox{\textbf{Junyang Lin}$^{2}$} \quad
    \mbox{\textbf{Jiangchao Yao}$^{1}$\thanks{Correspondence to: Jiangchao Yao
      (\texttt{sunarker@sjtu.edu.cn}).}}
  }\\[4pt]
  \parbox{\dimexpr\textwidth-2\tabcolsep\relax}{\centering\normalfont
    \mbox{$^{1}$CMIC, Shanghai Jiao Tong University} \quad
    \mbox{$^{2}$Qwen Team, Alibaba Group} \quad
    \mbox{$^{3}$Tongji University} \quad
    \mbox{$^{4}$Hong Kong Baptist University}
  }
}

\begin{document}

\maketitle

\begin{abstract}

% The emergence of the chain-of-thought paradigm further lengthens decoding in large language models (LLMs) and increases end-to-end latency. Speculative decoding (SD) alleviates this cost through a draft-then-verify paradigm, where a small draft model rapidly proposes candidate tokens and the target model verifies them in parallel. Prior work has mainly focused on reducing the cost of drafting and improving the alignment between draft and target models. As a result, the per-round verification cost has become a critical bottleneck to further speedups. Notably, compact variants within the same model family have demonstrated strong capabilities and remain well aligned with their larger counterparts due to shared training data and paradigms, making them effective tools for reducing verification cost. Motivated by these observations, we propose TriSpec, a ternary speculative decoding framework that integrates a lightweight proxy verifier to partially replace the target model in verification, and uses a margin-based routing mechanism to escalate uncertain tokens to the target for authoritative verification. Experiments on the Qwen3 and DeepSeek-R1-Distill-Qwen families show that TriSpec achieves up to 30\% additional speedup over standard SD, reduces target-model invocations by up to 50\%, and sustains accuracy with only minor degradation.

% The enhanced reasoning capabilities of Large Language Models (LLMs) have led to longer response sequences, yet the inference efficiency is fundamentally limited by their serial, autoregressive generation.
Inference efficiency in Large Language Models (LLMs) is fundamentally limited by their serial, autoregressive generation, especially as reasoning becomes a key capability and response sequences grow longer.
% Recent works have achieved notable acceleration through speculative decoding (SD), but mainly focus on draft model efficiency while neglecting the verification cost, an increasingly dominant factor. 
Speculative decoding (SD) offers a powerful solution, providing significant speed-ups through its lightweight drafting and parallel verification mechanism.
% While existing work has made considerable progress in improving the draft model's generation efficiency and alignment, this paper boosts SD from a new angle: the verification cost.
While existing work has nearly saturated improvements in draft effectiveness and efficiency, this paper advances SD from a new yet critical perspective: the verification cost.
% To address this issue, w
% We propose TriSpec, a novel ternary SD framework with proxy verification. In contrast to prior binary speculative decoding, TriSpec not only preserves its advantages but also alleviates the verification cost via a lightweight proxy verifier. 
% Specifically, the proxy performs low-cost pre-verification in place of the target model, and a margin-based routing policy decides whether to escalate to the target for authoritative verification. 
% We propose TriSpec, a novel ternary SD framework with proxy verification. At its core, TriSpec introduces a lightweight proxy model that handles the initial verification. This proxy significantly reduces computational cost by approving easily verifiable draft sequences and only engages the full target model when encountering uncertain tokens, thus striking an optimal balance between efficiency and quality.
We propose TriSpec, a novel ternary SD framework that, at its core, introduces a lightweight proxy to significantly reduce computational cost by approving easily verifiable draft sequences and engaging the full target model only when encountering uncertain tokens.
TriSpec can be integrated with state-of-the-art SD methods like EAGLE-3 to further reduce verification costs, achieving greater acceleration.
Extensive experiments on the Qwen3 and DeepSeek-R1-Distill-Qwen/LLaMA families show that TriSpec achieves up to 35\% speedup over standard SD, with up to 50\% fewer target model invocations while maintaining comparable accuracy.
Code is available at {\hypersetup{hidelinks}\href{https://github.com/haoyun-jiang/TriSpec}{\textcolor{blue!60!black}{\nolinkurl{https://github.com/haoyun-jiang/TriSpec}}}}.
\end{abstract}
\section{Introduction}
Large language models (LLMs)~\citep{qwen3,deepseekv3,llama3} have made remarkable success on complex tasks, with recent advances in reasoning leading to longer and more sophisticated responses.
This progress, however, is often hampered by the inherent latency of their serial, autoregressive generation process.
% remarkable progress on complex tasks, with their performance further boosted by recent advances in reasoning capability~\citep{deepseekr1,gpt4}. Meanwhile, the resulting increase in decoding length introduces significant inference cost, highlighting the necessity of reducing the cost and the end-to-end latency. 
% To \hf{address this efficiency bottleneck}, speculative decoding (SD)~\citep{sps} adopts a draft-then-verify paradigm, in which a small drafter rapidly proposes candidate tokens and the target model verifies them in parallel.
Speculative decoding (SD)~\citep{sps} addresses this efficiency bottleneck with a draft-then-verify paradigm. A small, fast drafter proposes candidate tokens, which the large target model then verifies in a single parallel pass, significantly accelerating inference without compromising output quality.
Generally, the overall speedup in speculative decoding is governed by three key factors: the speed of draft token generation, the acceptance rate of draft tokens, and the speed of verification. 
% (i) the speed of draft token generation, 
% (ii) the acceptance rate of draft tokens, 
% and (iii) the speed of verification. 
% \begin{enumerate}[topsep=0pt, itemsep=0pt, parsep=0pt,partopsep=0pt, leftmargin=35pt]
% \vspace{-3pt}
%     % \item \emph{Drafting efficiency: the overhead of generating draft tokens}
%     % \item \emph{Draft–target alignment: the acceptance rate of draft tokens;}
%     % \item \emph{Verification efficiency: the speed of the verification process.}
%     \item \emph{Drafting efficiency: draft generation speed}
%     \item \emph{Draft-target alignment: token acceptance rate}
%     \item \emph{Verification efficiency: validation speed}
% \vspace{-3pt}
% \end{enumerate}
Existing work has made substantial progress on the first two. 
Medusa~\citep{medusa}, EAGLE~\citep{eagle} and DFlash~\citep{dflash} propose lightweight drafter architectures, shortening the compute path and making the drafting time almost negligible. 
Building on this design, subsequent methods further improve acceptance rates by optimizing candidate paths~\citep{eagle2,adaeagle}, refining training paradigms~\citep{hass,eagle3,deepseekv3}, or relaxing verification constraints~\citep{judgedecoding,speculativecascade}, pushing this factor close to saturation. In contrast, the \emph{verification speed} remains an equally crucial yet largely overlooked factor. This gap motivates us to explore \emph{whether the verification role of the large target model could be handled by a more efficient, lightweight verifier}.

\begin{figure*}
    \centering
    \includegraphics[width=1\linewidth]{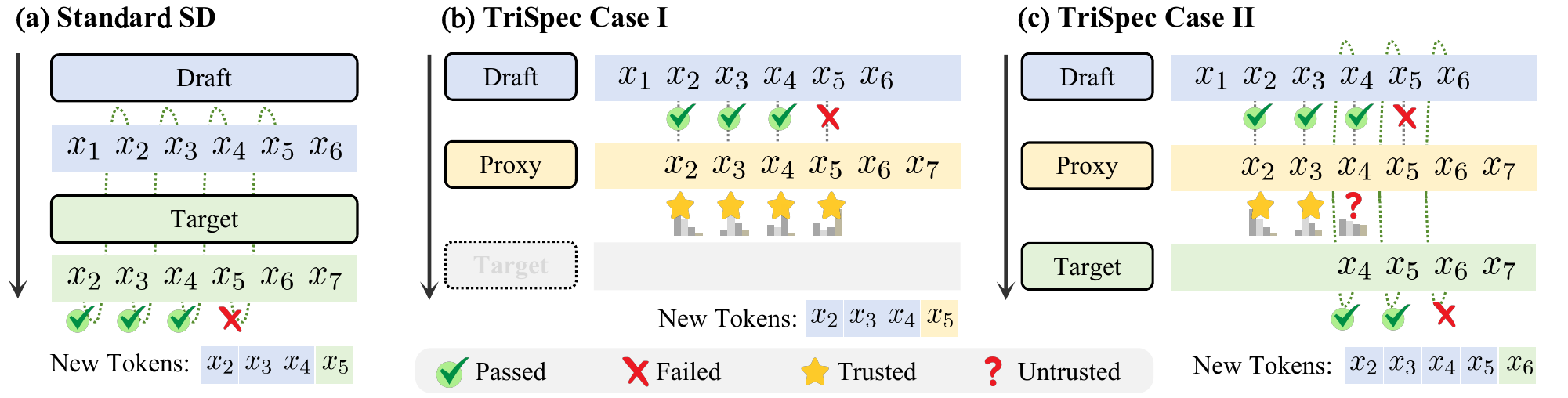}
    % \caption{Comparison of standard speculative decoding and TriSpec. Case I: The proxy verifier provides local correction of drafts, bypassing the large-scale target model to gain speed. Case II: If the proxy's verification is deemed untrusted before the first rejection, the corresponding tokens are escalated to the target model for authoritative validation to ensure accuracy. TriSpec employs a margin-based criterion to classify proxy verification as trusted or untrusted.}\
    \caption{Comparison of (a) standard SD and (b-c) TriSpec. In (b), the proxy remains trusted at the first rejection, enabling local correction without invoking the target. In (c), the proxy becomes untrusted before the first rejection, so the remaining draft tokens are escalated to the target for authoritative verification. A margin-based criterion determines whether proxy verification is trusted.}
    \vspace{-12pt}
    \label{fig:trispec}
\end{figure*}

We answer this question affirmatively. We find that smaller models from the same family as the target model (\emph{e.g.}, Qwen3-1.7B and Qwen3-32B) are promising candidates for lightweight verification, as their token-level predictions are highly consistent with those of the target model (\S\ref{sec:small_model}). Consequently, a small model can reliably replace the target model for verifying the majority of tokens, and the challenge lies in a small subset of tokens where their capabilities diverge. Fortunately, we find that a simple margin-based metric can sufficiently characterize this capability boundary, and route the verification task between a large target model and its smaller surrogate (Figure~\ref{fig:latency_breakdown}(b)).

% Many model families now offer compact variants (e.g., Qwen3-1.7B/0.6B and Gemma3-1B/0.27B) whose performance remains close to those of their larger counterparts despite having far fewer parameters~\citep{qwen3,gemma3}.
% Empirically, we find that these small models are implicitly aligned with the larger model, as they are trained on similar data using comparable training recipes.
% Moreover, because they retain nontrivial standalone reasoning ability, occasional deviations from the large model are tolerable~\citep{judgedecoding}.
% Given their compact size, these models impose substantially lower latency and memory footprint on each invocation, raising a natural question: Can compact variants replace the target model for portions of the verification process?

% To address this issue, 
Based on these insights, we propose TriSpec, a ternary speculative decoding paradigm that coordinates three models with complementary roles. A single-layer drafter provides high-throughput drafting; a lightweight proxy verifier quickly pre-verifies the draft and locally corrects a subset of tokens; and a large verifier supplies authoritative verification when needed (Figure~\ref{fig:trispec}). 
% Concretely, the drafter autoregressively grows a tree-structured draft. The proxy model then pre-verifies the drafted tokens, and a simple margin-based routing rule determines whether its verdict can be trusted for the current verification round. If the proxy cannot be trusted, the corresponding draft branch is pruned and the remaining tokens are escalated to the target for authoritative verification (see Figure~\ref{fig:trispec}).
% We evaluate TriSpec with Qwen3-32B and DeepSeek-R1-Distill-Qwen-32B as targets, using single-layer drafters trained via HASS~\citep{hass} or EAGLE3~\citep{eagle3}, and Qwen3-1.7B and DeepSeek-R1-Distill-Qwen-1.5B as assistant models. Across five standard benchmarks, TriSpec delivers significant speedups over binary speculative decoding with only limited accuracy loss. 
By delegating a substantial portion of verification to the cheaper proxy, TriSpec effectively reduces per-round verification time and thus the overall decoding latency.
Extensive experiments on Qwen3 and DeepSeek-R1-Distill-Qwen/LLaMA families demonstrate that TriSpec delivers significant speedups over SOTA speculative decoding methods.
In summary, our contributions are as follows:
\begin{itemize}
% [leftmargin=*]
\item We identify the verification cost as a new and critical perspective for accelerating speculative decoding, and show that smaller models within the same family exhibit strong alignment with their larger counterparts, making them effective and lightweight proxy verifiers.
% We propose a new perspective on reducing speculative decoding latency by lowering per-round verification time, and show that these smaller same-family models exhibit strong alignment with their larger counterparts, revealing their potential to serve as lightweight proxy verifiers.
% \item We propose TriSpec, a ternary speculative decoding paradigm that coordinates a drafter, a proxy verifier, and the target model. We design a margin-based routing rule that dynamically delegates verification, invoking the expensive target model only when necessary to ensure accuracy while minimizing overhead.
\item We introduce TriSpec, a ternary speculative decoding paradigm that jointly orchestrates a drafter, a proxy verifier, and the target model. A margin-based routing criterion adaptively allocates verification to the proxy for confident cases and invokes the expensive target only when necessary, preserving accuracy while substantially reducing verification overhead.
% \item We introduce TriSpec, a ternary speculative decoding paradigm that jointly orchestrates a drafter, a lightweight proxy verifier, and the target model. A margin-based routing rule adaptively allocates verification and invokes the expensive target only when necessary, preserving accuracy while substantially reducing verification overhead.
% coordinate a proxy and the target: the proxy performs fast pre-verification and calls the target only when necessary, thereby reducing verification overhead while preserving accuracy.
% \item Extensive experiments on widely used models and benchmarks show that TriSpec improves overall efficiency: it reduces target-model invocations by more than $2\times$, improves end-to-end speed by up to 30\% over standard SD, and maintains accuracy with an average drop no greater than 1\%.
% \item Extensive experiments show that TriSpec achieves up to a 30\% speedup over standard speculative decoding and reduces invocations of the target model by more than half. This efficiency is achieved with a minimal average accuracy drop of less than 1\%.
\item Extensive experiments on popular models and benchmarks show that TriSpec achieves up to a 35\% speedup over standard speculative decoding while reducing target-model invocations by more than half, with an average accuracy degradation of less than 1\%.
\end{itemize}
\section{Rethinking Speculative Decoding Speedup}

\subsection{Speculative Decoding}

% Standard speculative decoding follows a \emph{draft-then-verify} paradigm and involves two models: a draft model $M_d$ and a target model $M_t$~\cite{sps}. Let the vocabulary be denoted by $\mathcal{V}=\{1,2,\ldots, V\}$, and let $X \in \mathcal{V}^l$ represent the current decoding prefix with length $l$, including both the prompt and the previously generated tokens.
Standard speculative decoding follows a draft-then-verify paradigm with two models: a draft model $\mathcal{M}_d$ and a target model $\mathcal{M}_t$.
Let $\mathcal{V}$ denote a finite vocabulary with size $V \coloneqq |\mathcal{V}|$, and let the current decoding prefix be $X \in \mathcal{V}^{\,l}$, including both the prompt and previously generated tokens.

\begin{figure*}[t]
    \centering
    \includegraphics[width=1\linewidth]{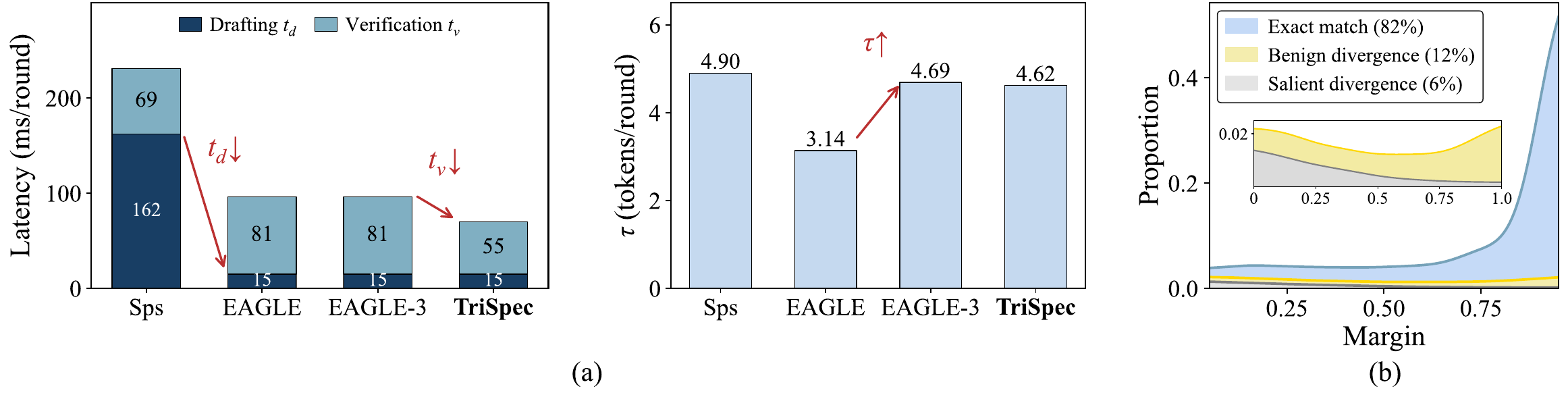}
    \vspace{-1em}
    \caption{(a) Latency decomposition of representative speculative decoding methods, with average per-round times shown on the left and acceptance lengths on the right. EAGLE reduces the drafting cost \(t_d\), EAGLE-3 improves the acceptance length \(\tau\), and TriSpec reduces the average verification cost \(t_v\). Measurements use Qwen3-32B on GSM8K with batch size 1, a single A100 GPU, and draft length 6. (b) Token outcomes versus the proxy's top-1--top-2 margin on ShareGPT. Exact matches agree with the target; mismatches passing the target-or-judge criterion (Appendix~\ref{sec:appendix-token-level-experiments}) are benign; others are salient. Salient divergences cluster at low margins. Stacked areas are globally normalized.}
    \label{fig:latency_breakdown}
\vspace{-12pt}
\end{figure*}

\noindent\textbf{Drafting phase.}
% During drafting, $\mathcal{M}_d$ autoregressively produces $k$ draft tokens and their probability distributions:
Conditioned on the current prefix $X$, the draft model $\mathcal{M}_d$ runs autoregressively for $k$ steps. At each step, it predicts a next-token distribution and samples one draft token:
\begin{equation}
\bigl(x^{(d)}_1, \mathbf{p}^{(d)}_1\bigr), \ldots, \bigl(x^{(d)}_k, \mathbf{p}^{(d)}_k\bigr) \leftarrow \mathcal{M}_d(X),
\label{eq:eq1}
\end{equation}
% where each $\mathbf{p}^{(d)}_i \in \mathbb{R}^{V}$ is the softmax probability distribution produced by the draft model at position $i$ and $x^{(d)}_i \in \mathcal{V}$ is sampled from $\mathbf{p}^{(d)}_i$.
% For a distribution $\mathbf{p}$ and token $x\in\mathcal{V}$, we write $\mathbf{p}(x)$ for the probability assigned to $x
where $\mathbf{p}^{(d)}_i \in \mathbb{R}^{V}$ is the draft distribution at position $i$, and $x^{(d)}_i \in \mathcal{V}$ is the sampled draft token. For any distribution $\mathbf{p}$ over $\mathcal{V}$ and token $x\in\mathcal{V}$, $\mathbf{p}(x)$ denotes the probability assigned to $x$.

% During the drafting stage, $M_d$ autoregressively generates $k$ probability distributions along with corresponding sampled tokens:
% \begin{equation}
% (x^{(d)}_1, p^{(d)}_1),\; (x^{(d)}_2, p^{(d)}_2),\; \ldots,\; (x^{(d)}_k, p^{(d)}_k) \;\leftarrow\; M_d(X),
% \label{eq:eq1}
% \end{equation}
% where $p^{(d)}_i \in \mathbb{R}^V$ is the softmax probability distribution produced by the draft model at position $i$, and $x^{(d)}_i \in \mathcal{V}$ is a token sampled from $p^{(d)}_i$. Throughout, for a probability vector $p$ and a token $x$, we write $p(x)$ to denote the probability of token $x$ under $p$.

\noindent\textbf{Verification phase.}
% In the verification stage, the target model $\mathcal{M}_t$ evaluates the $k$ drafted tokens in a single parallel forward pass:
The target model $\mathcal{M}_t$ verifies the $k$ drafted tokens with one parallel forward pass, producing target-side predictions for all draft positions and one bonus position:
\begin{equation}
\begin{aligned}
& (x^{(t)}_1, \mathbf{p}^{(t)}_1), (x^{(t)}_2, \mathbf{p}^{(t)}_2), \ldots, (x^{(t)}_{k+1}, \mathbf{p}^{(t)}_{k+1}) 
 \leftarrow \mathcal{M}_t(x^{(d)}_1, x^{(d)}_2, \ldots, x^{(d)}_k \mid X)
\end{aligned}
\end{equation}
Here, $\mathbf{p}^{(t)}_i \in \mathbb{R}^{V}$ is the target distribution at position $i$, and $x^{(t)}_i \in \mathcal{V}$ is the corresponding target token.
Each drafted token $x^{(d)}_i$ is accepted with probability $\min(1,\, \mathbf{p}^{(t)}_i(x^{(d)}_i)/\mathbf{p}^{(d)}_i(x^{(d)}_i))$.
If any rejection occurs, all drafted tokens to its right are discarded, and a replacement token is sampled from the normalized positive residual $\mathrm{norm}(\max(0, \mathbf{p}^{(t)}_i-\mathbf{p}^{(d)}_i))$ at the first rejection position.
Under this acceptance rule, the resulting samples follow exactly the same distribution as naive target-model decoding, which makes standard SD a provably lossless acceleration method~\cite{lossless}.

% \subsection{Latency Analysis of Speculative Decoding}

% \begin{lemma}[Latency decomposition for SD]\label{lemma: Latency decomposition for SD}
% Consider a speculative decoding process that generates $N$ tokens.
% Let $\tau$ denote the average acceptance length (tokens accepted per verification round), and let $t_d$, $t_v$, and $t_o$ be the average per-round drafting time, verification time, and other overhead, respectively.
% Then the end-to-end latency satisfies
% \begin{equation}
% L \;=\; \frac{N}{\tau}\bigl(t_d + t_v + t_o\bigr).
% \end{equation}
% In practice, $t_o$ is typically small and nearly constant compared to $t_d$ and $t_v$, and $N$ can be regarded as approximately constant for a given model.
% Consequently,
% \begin{equation}
% L \;\propto\; \frac{t_d + t_v}{\tau}.
% \end{equation}
% \end{lemma}
% \begin{conclusionbox}
% \textbf{Remark.} As modern speculative decoding methods substantially reduce $t_d$ and increase $\tau$, reducing the per-round verification time $t_v$ becomes an important and promising direction for further acceleration.
% \end{conclusionbox}

\subsection{Latency Analysis of Speculative Decoding}

\noindent\textbf{Latency decomposition for SD.}
Consider a speculative decoding process that generates $N$ tokens.
Let $\tau$ denote the average acceptance length, and let $t_d$, $t_v$, and $t_o$ denote the average per-round drafting time, verification time, and other overhead, respectively, computed over all rounds required to generate these $N$ tokens.
The end-to-end latency can be decomposed as
\begin{equation}
L = \frac{N}{\tau}(t_d+t_v+t_o)
\;\;\Longrightarrow\;\;
L \propto \frac{t_d+t_v}{\tau} \text{~at~} t_o\rightarrow 0.
\label{eq:latency_decomposition}
\end{equation}
% This simplification is well motivated in practice since $t_o$ is typically minor and stable across decoding rounds, while $N$ can be treated as approximately constant for a given model.
The proportional form assumes a fixed output length \(N\) and omits the minor, stable overhead \(t_o\), which is well motivated in practice under the same evaluation setting.

\begin{conclusionbox}
\textbf{New perspective.} As modern SD methods substantially reduce $t_d$ and increase $\tau$, reducing verification time $t_v$ becomes an important and promising direction for further acceleration.
\end{conclusionbox}

\textbf{Drafting has been largely optimized.}
Recent latency reductions in SD have mainly come from decreasing $t_d$ and increasing $\tau$.
Figure~\ref{fig:latency_breakdown}(a) illustrates how drafting time, verification time, and acceptance length evolve across representative SD methods.
Early SD methods~\cite{sps} often use a separate small model as the drafter, but its multi-layer architecture makes drafting latency-dominant.
Single-layer drafter designs, such as Medusa~\cite{medusa} and EAGLE~\cite{eagle3}, substantially reduce $t_d$ while maintaining a high $\tau$, making them the prevailing choice.
Subsequent methods further increase $\tau$ through tree-structured candidate paths and enhanced training schemes, as exemplified by EAGLE-3~\cite{eagle3}.
With Multi-Token Prediction (MTP)~\cite{deepseekv3} integrating draft-model training into pre-training, the acceptance rate of draft tokens can even reach up to 90\%.

% \textbf{Verification remains the dominant bottleneck.} Little progress has been made in directly reducing the per-round verification time \(t_v\). In state-of-the-art pipelines (e.g., EAGLE-3), verification time has become the dominant source of per-round latency, as each target-model invocation traverses a long computational path with a large parameter budget. A straightforward idea is to employ a lightweight verifier to offload part of the verification from the target, thereby reducing the average verification time across rounds.

\textbf{Verification remains the dominant bottleneck.}
In contrast, relatively little progress has been made in directly reducing the per-round verification time $t_v$.
As shown in the left panel of Figure~\ref{fig:latency_breakdown}(a), verification has become the dominant component of per-round latency in state-of-the-art pipelines such as EAGLE-3, since each target-model invocation must traverse a deep computation path with a large parameter budget.
This motivates a lightweight verifier that can offload part of the verification workload from the target, thereby reducing the average verification time across rounds.

\section{TriSpec}
\label{sec:TriSpec}
In this section, we propose TriSpec, a tightly-coupled ternary speculative decoding framework guided by a simple principle: \emph{use the cheapest component to fulfill each role}.
Specifically, TriSpec integrates a \emph{single-layer drafter} for low-cost token generation, a \emph{lightweight proxy verifier} to rapidly validate easy tokens and flag uncertain ones, and the \emph{target model} for exact verification of flagged cases to ensure correctness. By properly offloading most verification from the target to the proxy, TriSpec reduces the effective per-round verification time $t_v$ while maintaining accuracy.

% In this section, we propose TriSpec, a ternary speculative decoding framework that integrates a single-layer drafter (draft model), a lightweight proxy verifier (proxy), and the target model.
% The drafter leverages its single-layer architecture to produce candidates at very low latency, the lightweight proxy verifier pre-verifies drafts and locally corrects in certain cases, while the target model provides authoritative verification for uncertain tokens.

\subsection{Proxy choice: same-family smaller models}
\label{sec:small_model}

% \begin{wrapfigure}{r}{0.4\linewidth}  % r: 右侧放置；0.5\linewidth: 宽度
% \vspace{-1.2em}
%   \centering
%   \includegraphics[width=\linewidth]{Pictures/margin.pdf}
%   \caption{Distribution of token outcomes versus the proxy’s top-1–top-2 probability margin on ShareGPT. Areas are stacked and globally normalized, so the total across all classes sums to 1.}
%   \vspace{-1em}
%   \label{fig:margin_distribution}
% \end{wrapfigure}

% We identify same-family smaller models as ideal proxy verifiers, justified by the following three core properties.
% We choose same-family smaller models as proxy verifiers for their low inference cost and reliable token-level consistency with the target, as reflected by the following three properties.
Same-family smaller models are well suited as proxy verifiers due to their lower inference cost and inherent target alignment, as further evidenced by the following three properties.

\textbf{Strong Alignment.} 
Since speculative decoding verifies draft tokens position-wise, an effective proxy should closely align with the target’s token-level decisions. We evaluate this alignment on ShareGPT using Qwen3-family models by categorizing each next-token prediction as an exact match, benign divergence, or salient divergence. Benign divergences are alternative tokens deemed acceptable by the target or a stronger LLM, while the rest are treated as salient, indicating potential semantic deviation. Under this protocol, the proxy achieves an 82\% exact-match rate with the target, substantially higher than the EAGLE-3 drafter’s 68\%, with only 6\% salient divergences, suggesting that it can replace the target in selected verification cases. Details are provided in Appendix~\ref{sec:appendix-token-level-experiments}.

\textbf{Trustworthy outputs.}
% Beyond strong alignment with the target, the proxy’s outputs are themselves trustworthy. In the standalone example (Appendix~\ref{sec:appendix-standalone}), despite phrasing differences, the proxy and target reach the same final answer, indicating that modest token-level discrepancies between them are tolerable. Consistently, the end-to-end relaxed-verification experiments (Appendix~\ref{sec:appendix-relax-verification}) show that limited deviations from the target do not materially affect accuracy.
% Beyond token-level alignment, the proxy's outputs are also reliable at the semantic and task levels. In the standalone example in Appendix~\ref{sec:appendix-standalone}, the proxy differs from the target in wording and intermediate expressions, yet solves the problem correctly and reaches the same final answer, suggesting that modest proxy-target divergences can be semantically benign. Consistently, the end-to-end relaxed-verification experiments in Appendix~\ref{sec:appendix-relax-verification} show that limited high-confidence deviations from the target introduce little accuracy degradation.
Beyond token-level alignment, the proxy is reliable at semantic and task levels. Even when its outputs differ from the target in wording or intermediate steps, the proxy can still reach correct final answers, suggesting that a certain degree of divergence does not necessarily imply task-level failure (see Appendix~\ref{sec:appendix-standalone}). End-to-end relaxed-verification results further show that limited high-confidence proxy deviations introduce little accuracy degradation (see Appendix~\ref{sec:appendix-relax-verification}).

\textbf{Clear Separability.}
% A crucial property for an intermediate verifier is the existence of a clear indicator to distinguish between trusted and uncertain outputs. We find that the margin between the proxy's top-1 and top-2 probabilities serves this purpose effectively. As shown in Figure~\ref{fig:latency_breakdown}(b), this confidence margin cleanly stratifies tokens: a large margin (e.g., \(>0.5\)) correlates strongly with acceptable predictions, while unacceptable tokens are concentrated in low-margin cases. This property provides a robust and actionable criterion for assessing the reliability of the proxy's verification.
A useful proxy verifier should not only be aligned with the target, but also expose a simple signal to identify when its predictions are reliable. We find that the margin between the proxy's top-1 and top-2 probabilities serves this role. As shown in Figure~\ref{fig:latency_breakdown}(b), this confidence margin separates token outcomes: large-margin predictions are dominated by exact matches and benign divergences, whereas salient divergences concentrate in low-margin regions. This separability provides a practical and robust criterion for deciding when proxy verification can be trusted.

% For routing, the key question is which positions can be trusted under proxy verification. Intuitively, when the proxy is more confident in its top-1 prediction, it is less likely to diverge from the target, and its influence on the final answer is smaller. We quantify verification confidence as the margin between the proxy’s top-1 and top-2 probabilities at each position. As shown in Figure~\ref{fig:latency_breakdown}(b), this margin cleanly stratifies tokens: when the margin is large (e.g., \(>0.5\)), most proxy predictions are acceptable, whereas unacceptable tokens concentrate near zero. Thus, the margin provides an effective, actionable criterion for classifying the proxy’s output as trusted or untrusted.

% \subsection{Draft model training}
\subsection{Ternary speculative decoding}
\label{section:assistant_verification}
% \begin{wrapfigure}{r}{0.4\linewidth}  % r: 右侧放置；0.5\linewidth: 宽度
%   \centering
%   \includegraphics[width=\linewidth]{Pictures/draft_training.pdf}
%   \caption{Draft model training pipeline. }
%   \label{fig:drafter-training}
% \end{wrapfigure}

% \begin{figure*}[t!]
%     \centering
%     \includegraphics[width=1\linewidth]{Pictures/draft_training.pdf}
%     \caption{(a) Training pipeline of the draft model and adapter in TriSpec. The green, yellow, and gray modules represent the target, proxy, and drafter components, respectively.  (b) Draft tree pruning and tree attention mask mechanisms in TriSpec.}
%     \vspace{-1em}
%     \label{fig:method}
% \end{figure*}

\begin{figure}[t!]
    \centering
    \includegraphics[width=\linewidth]{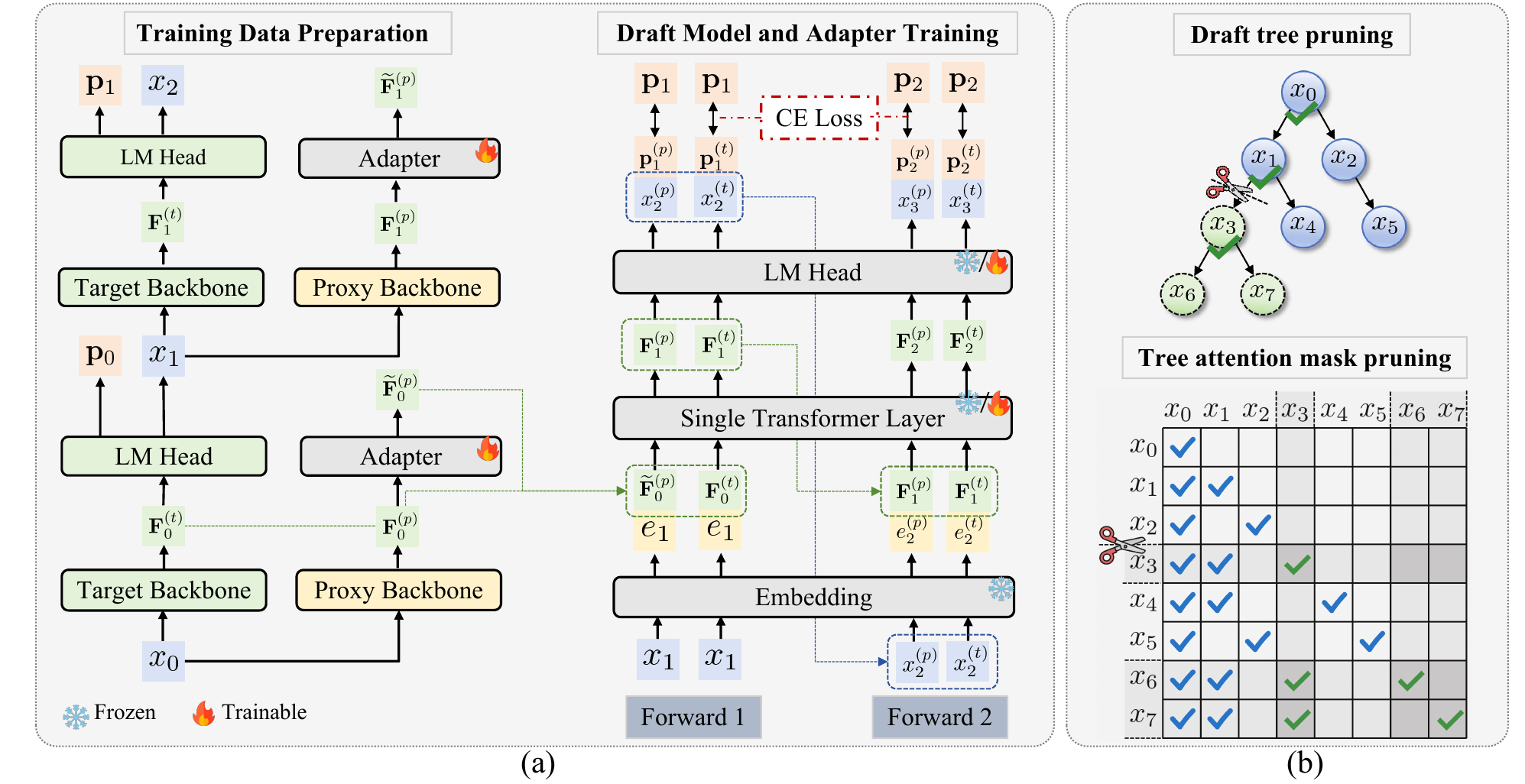}
    \caption{(a) Draft-model and adapter training in TriSpec. The adapter maps proxy features into the drafter's feature space; green, yellow, and gray modules denote target, proxy, and drafter components. (b) Token pruning and tree-attention mask updates after proxy-side verification in TriSpec.}
    \label{fig:draft_training}
    \vspace{-15pt}
\end{figure}

\textbf{Drafter selection and adapter training.}
% We adopt the EAGLE single-layer architecture as our draft model for its state-of-the-art efficiency. To support TriSpec's proxy-based verification, the drafter must flexibly process features from either the target or the proxy model. As in Figure~\ref{fig:draft_training}, we augment the EAGLE architecture with a one-layer MLP adapter that maps proxy features into the drafter’s expected feature space. This architecture supports two training regimes: (i) \textit{joint training}, where the drafter and adapter are optimized together using both proxy and target features; and (ii) \textit{adapter-only finetuning}, where a pretrained drafter's weights are fixed and only the adapter is trained using proxy features. We compare these regimes in Section~\ref{sec:draft-training}.
We adopt the EAGLE single-layer architecture as our drafter for its state-of-the-art efficiency. Since EAGLE-style drafters are conditioned on target-model features, TriSpec further requires them to process proxy-side features. As shown in Figure~\ref{fig:draft_training}(a), we add a one-layer MLP adapter to map proxy features into the drafter's feature space. This design supports two training regimes: (i) \textit{joint training}, where the drafter and adapter are optimized together with target and proxy features; and (ii) \textit{adapter-only finetuning}, where the pretrained drafter is frozen and only the adapter is trained on proxy features. We compare them in Section~\ref{sec:draft-training}.

% Because the assistant and target models are strongly aligned, the assistant can pre-validate drafter proposals before invoking the target.
% Given the lower query cost of the proxy verifier, we first use it to pre-validate the draft tokens before invoking the target. With the \(k\) drafted pairs \((x^{(d)}_1,p^{(d)}_1),\ldots,(x^{(d)}_k,p^{(d)}_k)\) in Eq.~\ref{eq:eq1}, the proxy verifier \(M_a\) performs parallel validation of these drafts:
% \begin{equation}
% (x^{(a)}_1, p^{(a)}_1),\; (x^{(a)}_2, p^{(a)}_2),\; \ldots,\; (x^{(a)}_{k+1}, p^{(a)}_{k+1})
% \;\leftarrow\; M_a(x^{(d)}_1, x^{(d)}_2, \ldots, x^{(d)}_k \mid X),
% \end{equation}
\textbf{Proxy pre-verification.}
% Given the drafted pairs \((x^{(d)}_1,\mathbf{p}^{(d)}_1),\ldots,(x^{(d)}_k,\mathbf{p}^{(d)}_k)\), we leverage the proxy \(\mathcal{M}_p\) to perform a single parallel pass and returns its own token predictions and distributions:
Given the drafted pairs \((x^{(d)}_1,\mathbf{p}^{(d)}_1),\ldots,(x^{(d)}_k,\mathbf{p}^{(d)}_k)\), we first invoke the proxy verifier \(\mathcal{M}_p\) to pre-verify the draft and obtain proxy-side predictions and distributions:
\begin{equation}
\begin{aligned}
& \bigl(x^{(p)}_1, \mathbf{p}^{(p)}_1\bigr), \bigl(x^{(p)}_2, \mathbf{p}^{(p)}_2\bigr), \ldots, \bigl(x^{(p)}_{k+1}, \mathbf{p}^{(p)}_{k+1}\bigr) \leftarrow \mathcal{M}_p \bigl(x^{(d)}_1, x^{(d)}_2, \ldots, x^{(d)}_k \,\big|\, X\bigr),
\end{aligned}
\end{equation}
% where \(x^{(p)}_i\in\mathcal{V}\) and \(\mathbf{p}^{(p)}_i\in\mathbb{R}^{V}\) are the proxy’s predicted token and its softmax distribution.
% Following standard speculative decoding, the proxy determines acceptance by comparing its probability distribution with that of the drafter. Formally, for each draft token $x^{(d)}_i$ where $i \in \{1, \ldots, k\}$, we define a Bernoulli acceptance variable
where \(x^{(p)}_i\in\mathcal{V}\) and \(\mathbf{p}^{(p)}_i\in\mathbb{R}^{V}\) denote the proxy token and distribution at position \(i\).
Following standard speculative decoding, the proxy accepts each draft token by comparing its probability with the drafter's. Formally, for \(i \in \{1,\ldots,k\}\), we define a Bernoulli acceptance variable:
\begin{equation}
s_i \sim \mathrm{Bernoulli} \left( \min \left\{ 1, \frac{\mathbf{p}^{(p)}_i(x^{(d)}_i)}{\mathbf{p}^{(d)}_i(x^{(d)}_i)} \right\} \right),
\label{eq:acceptance_prob}
\end{equation}
which in the greedy case reduces to $s_i = \mathbf{1}\{x^{(p)}_i = x^{(d)}_i\}$. 
Then, the proxy-side acceptance length is
\begin{equation}
\tau_a \;=\; \min \big( \{\, i \;\big|\; 0 \le i \le k-1,\; s_{i+1} = 0 \} \,\cup\, \{k\} \big).
\end{equation}
\textbf{Margin-based Criterion.}
% To preserve predictive integrity, TriSpec identifies proxy-target divergence using the confidence margin between the proxy's top-1 and top-2 probabilities, which reflects the reliability of proxy verification as motivated in \S\ref{sec:small_model}. We thus define a margin-based binary predicate \(g: \mathbb{R}^{V} \rightarrow \{0,1\}\) to determine whether a proxy verification result can be trusted:
To avoid unreliable proxy decisions, TriSpec uses the confidence margin between the proxy's top-1 and top-2 probabilities as a reliability signal, as motivated in Sec.~\ref{sec:small_model}. We define a margin-based predicate \(g: \mathbb{R}^{V} \rightarrow \{0,1\}\) to indicate whether a proxy prediction is trusted:
\begin{equation}
g\big(\mathbf{p}^{(p)}_i\big) = \mathbf{1}\left\{ \operatorname{top}_1\big(\mathbf{p}^{(p)}_i\big) - \operatorname{top}_2\big(\mathbf{p}^{(p)}_i\big) \ge \lambda \right\},
\label{eq:margin_predicate}
\end{equation}
% where $i \in \{1, \ldots, k+1\}$, $\operatorname{top}_1(x)$ and $\operatorname{top}_2(x)$ denote the largest and second-largest entries of $x$, and $\lambda\in(0,1)$ is a tunable threshold controlling the accuracy–latency trade-off. To measure how long we can rely on the proxy before escalation, we define the proxy-trusted prefix length as the longest initial segment that passes the margin test:
where \(i \in \{1,\ldots,k+1\}\), \(\operatorname{top}_1(x)\) and \(\operatorname{top}_2(x)\) denote the largest and second-largest entries of \(x\), and \(\lambda\in(0,1)\) controls the accuracy--latency trade-off. Based on this predicate, we define the proxy-trusted prefix length, which measures how long proxy verification remains reliable:
% \begin{equation}
% \tau_m \;=\; \min \big( \{\, i \;\big|\; 0 \le i \le k,\; g(\mathbf{p}^{(p)}_i+1)=0 \} \,\cup\, \{k+1\} \big).
% \end{equation}
\begin{equation}
\tau_m = \min \bigl( \{ i \mid 0 \le i \le k, g(\mathbf{p}^{(p)}_{i+1})=0 \} \cup \{k+1\} \bigr).
\label{eq:tau_m}
\end{equation}
% \begin{equation}
% \begin{aligned}
% \tau_m = \min \bigl( \{ i-1 \mid \, & 1 \le i \le k+1, g(\mathbf{p}^{(p)}_i)=0 \} \\
% & \cup \{k+1\} \bigr).
% \end{aligned}
% \end{equation}
% with $\tau_m = k+1$ if all $g(\mathbf{p}^{(p)}_i)=1$, meaning the proxy’s verification is trusted for the entire draft and it can reliably generate the $(k+1)$-th token.
In particular, if $g(\mathbf{p}^{(p)}_i)=1$ for all $i$, then $\tau_m = k+1$, indicating that the proxy’s verification is trusted for the entire draft and can reliably provide the $(k+1)$-th token.

% \begin{figure}[t!]
%     \centering
%     % \vspace{-10pt}
%     % \includegraphics[width=1\linewidth]{Pictures/tree_pruning1.pdf}
%     \includegraphics[width=0.9\linewidth]{Pictures/fig4_icml.pdf}
%     \caption{Token pruning and attention mask updates in TriSpec.}
%     \label{fig:tree_pruning}
%     \vspace{-10pt}
% \end{figure}

% Accordingly, we define the corresponding confidence mask \(m_j\in\{0,1\}^k\) by
% \begin{equation}
% m_j[i] \;=\; \prod_{t=1}^{i} g(r_{l+t}),\quad i=1,\ldots,k,
% \end{equation}
% Let \(\tau_a=\sum_{i=1}^{k} m_a[i]\) be the number of consecutively accepted tokens under the assistant’s verification rule and \(\tau_m=\sum_{i=1}^{k} m_j[i]\) the longest prefix on which the assistant is confident.

% \textbf{Routing.}

% \input{Algorithm/algorithm1}

% Then, the procedure of verification can be divided into two cases, depending on the necessity of invoking the target: 

% \input{Tables/table1_mainresult}

\textbf{Adaptive Verification Routing.}
Based on the proxy-side acceptance length \(\tau_a\) and the proxy-trusted prefix length \(\tau_m\), TriSpec adaptively routes verification between the proxy and target into two cases:

(i) If \(\tau_a < \tau_m\), the proxy remains trusted at the first rejection point and completes the round without invoking the target model (Figure~\ref{fig:trispec}(b)). Accordingly, we accept \(x^{(d)}_{1:\tau_a}\) and locally correct the next token with \(x^{(p)}_{\tau_a+1}\). The generated sequence for this round is \(Y = \{x^{(d)}_1,\ldots,x^{(d)}_{\tau_a},x^{(p)}_{\tau_a+1}\}\).

% (ii) If \(\tau_a \ge \tau_m\), the proxy’s verification is deemed untrusted before the first rejection point, so the verification is thus delegated to the target model (Figure~\ref{fig:trispec} right). We first accept \(x^{(d)}_{1:\tau_m}\), prune these tokens from the draft tree (Figure~\ref{fig:draft_training}(b)), and pass the remaining branches to the target for validation:
(ii) If \(\tau_a \ge \tau_m\), the proxy verification becomes untrusted before reaching the first rejection point, so verification is delegated to the target model (Figure~\ref{fig:trispec}(c)). We first accept \(x^{(d)}_{1:\tau_m}\), prune these tokens from the draft tree (Figure~\ref{fig:draft_training}(b)), and pass the remaining branches to the target for validation:
\begin{equation}
\begin{aligned}
& (x^{(t)}_{\tau_m+1}, \mathbf{p}^{(t)}_{\tau_m+1}), \ldots, (x^{(t)}_{k+1}, \mathbf{p}^{(t)}_{k+1})
 \leftarrow \mathcal{M}_t \bigl(x^{(d)}_{\tau_m+1}, \ldots, x^{(d)}_k \mid X, x^{(d)}_{1:\tau_m}\bigr),
\end{aligned}
\end{equation}
Applying the standard left-to-right acceptance rule to \(\mathbf{p}^{(t)}\) gives the target-verified acceptance length \(\tau_t\). The generated sequence for this round is $Y = \{x^{(d)}_1,\ldots,x^{(d)}_{\tau_m+\tau_t},x^{(t)}_{\tau_m+\tau_t+1}\}$.

\subsection{Theoretical Error Analysis of TriSpec}
\label{sec:error_analysis}

TriSpec allows the proxy verifier to finalize selected token decisions, introducing possible deviations from target-only decoding. We analyze their effects on task-level performance and the sequence-level output distribution.

\paragraph{Task-level Performance Gap.}
Let $\Delta_A=\mathbb E[U_t]-\mathbb E[U_A]$ be the signed gap in normalized task score $U_A\in[0,1]$, with expectations over inputs and decoding randomness. On a common target-only reference trajectory, a position is task-critical if a single-token intervention using a fixed potential proxy token changes the final score under coupled subsequent randomness. Over pooled reference task-critical positions, let $F_\lambda$ denote proxy finalization and $S$ a salient divergence, a surrogate for potentially harmful decisions. Define $p_\lambda=\Pr(F_\lambda)$, $\mu_0=\Pr(S)$, and, when $p_\lambda>0$, $\mu_\lambda=\Pr(S\mid F_\lambda)$. For $p_\lambda>0$ and $\mu_0>0$, assuming comparable average signed contributions from salient divergences and neglecting their interactions and subsequent trajectory changes, we obtain the first-order estimate
\begin{equation}
\label{eq:error_estimate}
\Delta_{\mathrm{tri}}
\approx
p_\lambda
\frac{\mu_\lambda}{\mu_0}
\Delta_{\mathrm{proxy}} .
\end{equation}
Under these conditions, the scaling factor is exactly $\rho_\lambda=\Pr(F_\lambda\mid S)$: the fraction of salient divergences finalized by the proxy among reference task-critical positions. The performance-gap relation remains a first-order approximation; when $\Delta_{\mathrm{proxy}}>0$ and $\rho_\lambda<1$, it predicts reduced degradation without guaranteeing it. Figure~\ref{fig:latency_breakdown}(b) provides qualitative support for margin filtering, not estimates of these latent probabilities. Appendix~\ref{app:task_gap} specifies the reference construction, approximation assumptions, derivation, and boundary cases.

\paragraph{Sequence-level Distributional Deviation.}
Task-level performance preservation does not imply distributional equivalence. For a fixed prompt and sampling configuration, let $P_\lambda^{(T)}$ and $P_t^{(T)}$ denote the distributions of $T$-token continuations under TriSpec and target-only decoding. Let $\mathbb E[N_F(\lambda)]$ count proxy-finalized tokens in expectation, $\varepsilon_\lambda$ be their average conditional KL divergence from the target, and $B_\lambda$ be the accumulated target-routed conditional KL contribution, including routing selection effects. The KL chain rule, convexity, and Pinsker's inequality give
\begin{equation}
\label{eq:sequence_deviation}
\begin{aligned}
\mathrm{KL}\!\left(P_\lambda^{(T)}\middle\|P_t^{(T)}\right)
&\leq B_\lambda+\mathbb E[N_F(\lambda)]\varepsilon_\lambda,\\
\mathrm{TV}\!\left(P_\lambda^{(T)},P_t^{(T)}\right)
&\leq \sqrt{\frac{B_\lambda+\mathbb E[N_F(\lambda)]\varepsilon_\lambda}{2}}.
\end{aligned}
\end{equation}
These bounds characterize the sources of deviation; without estimating their terms, they do not establish small actual bias. The simplification $B_\lambda=0$ additionally requires target-routed conditionals to match the target distribution. Increasing $\lambda$ does not guarantee a decreasing bound, and KL may be infinite under greedy decoding. Appendix~\ref{app:sequence_bound} gives the precise definitions, proof, support conditions, and zero-exposure convention.

\subsection{Further Discussion}

Unlike Judge Decoding~\cite{judgedecoding} and SpecCascade~\cite{speculativecascade}, which increase the average acceptance length \(\tau\) by accepting misaligned yet high-quality tokens, TriSpec directly reduces verification time \(t_v\) by delegating easy cases to the proxy and avoiding unnecessary target calls. This is especially useful for target-feature-dependent drafters such as EAGLE-3 and MTP, where longer draft horizons often degrade prediction quality and limit further gains in \(\tau\). By using the proxy to verify and correct draft tokens, TriSpec alleviates this bottleneck and remains compatible with diverse drafter architectures. Moreover, unlike cascaded frameworks such as TriForce~\cite{triforce}, TriSpec better coordinates all components within a unified decoding loop to exploit their strengths and reduce redundant calls.

% The next drafting round then starts from the updated prefix \(T_{1:l+\tau_m+\tau_t+1}\):
% \begin{equation}
% (c_{l+\tau_m+\tau_t+2}, p_{l+\tau_m+\tau_t+2}),\ldots,(c_{l+\tau_m+\tau_t+k+1}, p_{l+\tau_m+\tau_t+k+1})
% \;\leftarrow\; M_d(T_{1:l+\tau_m+\tau_t+1};\,k).
% \end{equation}

\section{Experiments}
\label{sec:Experiments}
\begin{table*}[t!]
\centering
\caption{Accuracy and speedup on five reasoning benchmarks at temperature 0. ``SC'' denotes SpecCascade. In the Avg Acc. column, the small value denotes the change from target-only decoding.}
\scriptsize
\setlength{\tabcolsep}{2pt}  
\begin{tabular}{clcccccccccccc}
\toprule
& & \multicolumn{2}{c}{GSM8K} & \multicolumn{2}{c}{MATH500} & \multicolumn{2}{c}{GK23-en} & \multicolumn{2}{c}{HumanEval} & \multicolumn{2}{c}{MBPP} & \multicolumn{2}{c}{Avg} \\
\midrule
Model & \multicolumn{1}{c}{Method} & Acc. & Speedup & Acc. & Speedup & Acc. & Speedup & Acc. & Speedup & Acc. & Speedup & Acc. & Speedup \\ 
\midrule
\multirow{13}{*}{Qwen3} 
& Single Model (Target,32B)     & 98\% & 1.00$\times$ & 85\% & 1.00$\times$ & 75\% & 1.00$\times$ & 87\% & 1.00$\times$ & 77\% & 1.00$\times$ & 84.4\% & 1.00$\times$ \\ 
& Single Model (Proxy,1.7B)        & 86\% & 2.36$\times$ & 81\% & 3.01$\times$ & 70\% & 3.03$\times$ & 81\% & 2.93$\times$ & 51\% & 2.92$\times$ & 73.8\% {\tiny-10.6\%} & 2.85$\times$ \\
\cmidrule(lr){2-14}
& HASS     & 98\% & 2.98$\times$ & 86\% & 2.27$\times$ & 74\% & 2.30$\times$ & 87\% & 2.03$\times$ & 77\% & 2.18$\times$ & 84.4\% {\tiny$\pm$0\%} & 2.35$\times$\\
& HASS + SC[Chow]  & 80\% & 3.33$\times$ & 55\% & 2.60$\times$ & 51\% & 2.64$\times$ & 74\% & 2.23$\times$ & 56\% & 2.33$\times$ & 63.2\% {\tiny-21.2\%} & 2.63$\times$ \\
& HASS + SC[OPT] & 76\% & 3.25$\times$ & 38\% & 2.66$\times$ & 33\% & 2.65$\times$ & 49\% & 2.30$\times$ & 20\% & 2.34$\times$ & 43.2\%{\tiny-41.2\%} & 2.64$\times$ \\
& HASS + SC[Token]  & 98\% & 3.10$\times$ & 86\% & 2.34$\times$ & 74\% & 2.39$\times$ & 87\% & 2.15$\times$ & 74\% & 2.28$\times$ & 83.8\% {\tiny-0.6\%} & 2.63$\times$ \\
& \tscell{HASS + TriSpec} & \tscell{97\%} & \tscell{\textbf{3.85$\times$}} & \tscell{86\%} & \tscell{\textbf{3.18$\times$}} & \tscell{75\%} & \tscell{\textbf{3.26$\times$}} & \tscell{86\%} & \tscell{\textbf{2.99$\times$}} & \tscell{74\%} & \tscell{\textbf{2.80$\times$}} & \tscell{83.6\% {\tiny-0.8\%}} & \tscell{\textbf{3.22$\times$}}\\
\cmidrule(lr){2-14}
& EAGLE3     & 98\% & 3.25$\times$ & 87\% & 2.73$\times$ & 74\% & 2.77$\times$ & 87\% & 2.71$\times$ & 76\% & 2.77$\times$ & 84.4\% {\tiny$\pm$0\%} & 2.85$\times$ \\
& EAGLE3 + SC[Chow]    & 95\% & 3.39$\times$ & 70\% & 2.91$\times$ & 58\% & 2.94$\times$ & 76\% & 2.82$\times$ & 67\% & 2.85$\times$ & 73.2\% {\tiny-11.2\%} & 2.98$\times$ \\
& EAGLE3 + SC[OPT]  & 93\% & 3.43$\times$ & 58\% & 3.06$\times$ & 42\% & 3.01$\times$ & 61\% & 2.91$\times$ & 49\% & 2.87$\times$ & 60.6\%{\tiny-23.8\%} & 3.06$\times$ \\
& EAGLE3 + SC[Token]  & 98\% & 3.34$\times$ & 86\% & 2.90$\times$ & 74\% & 2.91$\times$ & 87\% & 2.78$\times$ & 76\% & 2.85$\times$ & 84.2\% {\tiny-0.2\%} & 2.96$\times$ \\
& \tscell{EAGLE3 + TriSpec} & \tscell{98\%} & \tscell{\textbf{4.18$\times$}} & \tscell{88\%} & \tscell{\textbf{3.49$\times$}} & \tscell{74\%} & \tscell{\textbf{3.55$\times$}} & \tscell{87\%} & \tscell{\textbf{3.27$\times$}} & \tscell{74\%} & \tscell{\textbf{3.24$\times$}} & \tscell{84.2\% {\tiny-0.2\%}} & \tscell{\textbf{3.55$\times$}}\\
\midrule
\multirow{13}{*}{DSQ}
& Single Model (Target,32B)        & 93\% & 1.00$\times$ & 86\% & 1.00$\times$ & 74\% & 1.00$\times$ & 85\% & 1.00$\times$ & 72\% & 1.00$\times$ & 82.0\% & 1.00$\times$ \\
& Single Model (Proxy,1.5B)  & 72\% & 3.09$\times$ & 76\% & 3.32$\times$ & 62\% & 3.33$\times$ & 54\% & 3.22$\times$ & 38\% & 3.32$\times$ & 60.4\% {\tiny-21.6\%} & 3.26$\times$ \\
\cmidrule(lr){2-14}
& HASS      & 93\% & 3.28$\times$ & 88\% & 2.99$\times$ & 72\% & 2.90$\times$ & 84\% & 2.78$\times$ & 73\% & 3.02$\times$ & 82.0\% {\tiny$\pm$0\%} & 2.99$\times$ \\ 
& HASS + SC[Chow]    & 77\% & 3.46$\times$ & 51\% & 3.22$\times$ & 44\% & 3.06$\times$ & 75\% & 2.85$\times$ & 62\% & 3.10$\times$ & 61.8\% {\tiny-20.2\%} & 3.14$\times$ \\
& HASS + SC[OPT] & 77\% & 3.59$\times$ & 39\% & 3.20$\times$ & 20\% & 3.18$\times$ & 41\% & 2.88$\times$ & 18\% & 3.12$\times$ & 39.0\%{\tiny-43.0\%} & 3.19$\times$ \\
& HASS + SC[Token]  & 91\% & 3.41$\times$ & 85\% & 3.13$\times$ & 69\% & 2.96$\times$ & 84\% & 2.80$\times$ & 72\% & 3.04$\times$ & 80.2\% {\tiny-1.8\%} & 3.07$\times$ \\
& \tscell{HASS + TriSpec} & \tscell{91\%} & \tscell{\textbf{3.76$\times$}} & \tscell{87\%} & \tscell{\textbf{3.41$\times$}} & \tscell{74\%} & \tscell{\textbf{3.33$\times$}} & \tscell{85\%} & \tscell{\textbf{2.98$\times$}} & \tscell{71\%} & \tscell{\textbf{3.24$\times$}} & \tscell{81.6\% {\tiny-0.4\%}} & \tscell{\textbf{3.34$\times$}} \\
\cmidrule(lr){2-14}
& EAGLE3    & 93\% & 3.42$\times$ & 88\% & 3.19$\times$ & 74\% & 3.11$\times$ & 86\% & 3.09$\times$ & 72\% & 3.27$\times$ & 82.6\% {\tiny+0.6\%} & 3.22$\times$ \\
& EAGLE3 + SC[Chow]    & 84\% & 3.67$\times$ & 69\% & 3.47$\times$ & 51\% & 3.43$\times$ & 72\% & 3.19$\times$ & 64\% & 3.33$\times$ & 68.0\% {\tiny-14.0\%} & 3.42$\times$ \\
& EAGLE3 + SC[OPT]  & 81\% & 3.69$\times$ & 58\% & 3.51$\times$ & 45\% & 3.46$\times$ & 54\% & 3.23$\times$ & 48\% & 3.33$\times$ & 57.2\%{\tiny-24.8\%} & 3.44$\times$ \\
& EAGLE3 + SC[Token]  & 92\% & 3.54$\times$ & 82\% & 3.36$\times$ & 72\% & 3.26$\times$ & 86\% & 3.18$\times$ & 70\% & 3.27$\times$ & 80.4\% {\tiny-1.6\%} & 3.32$\times$ \\
& \tscell{EAGLE3 + TriSpec} & \tscell{91\%} & \tscell{\textbf{4.14$\times$}} & \tscell{88\%} & \tscell{\textbf{3.69$\times$}} & \tscell{75\%} & \tscell{\textbf{3.63$\times$}} & \tscell{88\%} & \tscell{\textbf{3.32$\times$}} & \tscell{69\%} & \tscell{\textbf{3.57$\times$}} & \tscell{82.2\% {\tiny+0.2\%}} & \tscell{\textbf{3.67$\times$}} \\
\midrule
\multirow{8}{*}{DSL}
& Single Model (Target,70B)        & 88\% & 1.00$\times$ & 90\% & 1.00$\times$ & 78\% & 1.00$\times$ & 87\% & 1.00$\times$ & 79\% & 1.00$\times$ & 84.4\% & 1.00$\times$ \\ 
& Single Model (Proxy,8B)  & 77\% & 3.71$\times$ & 83\% & 3.73$\times$ & 73\% & 4.01$\times$ & 80\% & 3.89$\times$ & 59\% & 4.03$\times$ & 74.4\% {\tiny-10.0\%} & 3.87$\times$ \\
\cmidrule(lr){2-14}
& EAGLE3    & 88\% & 2.83$\times$ & 89\% & 2.35$\times$ & 76\% & 2.31$\times$ & 84\% & 2.49$\times$ & 80\% & 2.40$\times$ & 83.4\% {\tiny-1.0\%} & 2.48$\times$ \\
& EAGLE3 + SC[Chow]  & 82\% & 2.90$\times$ & 81\% & 2.50$\times$ & 65\% & 2.61$\times$ & 74\% & 2.58$\times$ & 71\% & 2.57$\times$ & 74.6\% {\tiny-9.8\%} & 2.63$\times$ \\
& EAGLE3 + SC[OPT]  & 78\% & 3.02$\times$ & 66\% & 2.84$\times$ & 48\% & 2.93$\times$ & 60\% & 2.89$\times$ & 51\% & 2.91$\times$ & 60.6\% {\tiny-23.8\%} & 2.92$\times$ \\
& EAGLE3 + SC[Token]  & 87\% & 2.92$\times$ & 88\% & 2.43$\times$ & 75\% & 2.56$\times$ & 85\% & 2.69$\times$ & 78\% & 2.65$\times$ & 82.6\% {\tiny-1.8\%} & 2.65$\times$ \\
& \tscell{EAGLE3 + TriSpec} & \tscell{86\%} & \tscell{\textbf{3.74$\times$}} & \tscell{88\%} & \tscell{\textbf{3.35$\times$}} & \tscell{77\%} & \tscell{\textbf{3.36$\times$}} & \tscell{87\%} & \tscell{\textbf{3.07$\times$}} & \tscell{76\%} & \tscell{\textbf{3.29$\times$}} & \tscell{82.8\% {\tiny-1.6\%}} & \tscell{\textbf{3.36$\times$}} \\
\bottomrule
\end{tabular}
\label{tab:model_performance}
\vspace{-5pt}
\end{table*}

% \begin{wrapfigure}{r}{0.5\linewidth}  % r: 右侧放置；0.5\linewidth: 宽度
%   \centering
%   \includegraphics[width=\linewidth]{Pictures/tree_pruning.pdf}
%   \caption{Draft tree pruning and tree attention modifications}
%   \label{fig:tree_pruning}
% \end{wrapfigure}

\subsection{Setup}
\label{sec:setup}
\textbf{Models and baselines.}
% We evaluate TriSpec across three model families: Qwen3, DeepSeek-R1-Distill-Qwen (DSQ), and DeepSeek-R1-Distill-LLaMA (DSL). We pair target models (Qwen3-32B, DSQ-32B, DSL-70B) with their same-family smaller versions as proxies (1.7B, 1.5B, 8B), using corresponding HASS/EAGLE3 weights for drafting. Baselines include standard speculative decoding and SpecCascade~\cite{speculativecascade} variants. All drafters are retrained on unified data for fair comparison. Following EAGLE-2 configuration, we set the draft tree depth 6, the expansion top-k 10, and the total draft budget tokens 60. For TriSpec, we employ the margin threshold $\lambda=0.5$. Further details are provided in Appendix~\ref{sec:appedix-draft-training} and~\ref{sec:appedix-baseline-detail}.
We evaluate TriSpec on three model families: Qwen3, DeepSeek-R1-Distill-Qwen (DSQ), and DeepSeek-R1-Distill-LLaMA (DSL). Target models (Qwen3-32B, DSQ-32B, DSL-70B) are paired with same-family smaller proxies (1.7B, 1.5B, 8B) and equipped with corresponding HASS/EAGLE3 drafters. Baselines include standard SD and SpecCascade~\cite{speculativecascade} variants. For a fair comparison, all drafters are retrained on the same data and, at inference, use tree depth 6, expansion top-$k$ 10, and a total draft budget of 60 tokens. Unless otherwise stated, TriSpec uses a margin threshold of $\lambda=0.5$. Further details are provided in Appendix~\ref{sec:appedix-draft-training} and~\ref{sec:appendix-baseline-detail}.

% \textbf{Drafter Training.}
% To ensure fairness, all draft models (both the baselines and TriSpec) are trained from scratch on the same data using the official HASS and EAGLE3 codebases. Following the respective papers, HASS is trained on 68K ShareGPT examples, and EAGLE3 additionally incorporates UltraChat-200K~\cite{ultrachat}. 
% % Concretely, we first generate responses autoregressively with the target model and then compute token-level hidden features offline with both the proxy and the target to supervise baseline training and our adapter.
% In the main experiments, the drafter’s backbone and adapter are jointly trained from scratch for TriSpec. Training is performed on 8$\times$ NVIDIA A100 80GB GPUs, with hyperparameters set according to the official recommendations unless otherwise noted. Other details are shown in Appendix.

% \input{Tables/table_temperature1}

\textbf{Benchmarks.}
We primarily evaluate our method on five reasoning benchmarks: three mathematical reasoning datasets (GSM8K~\cite{gsm8k}, MATH500~\cite{math500}, Gaokao-2023-EN) and two code-generation datasets (HumanEval~\cite{humaneval}, MBPP~\cite{mbpp}). To further assess performance across diverse domains and more challenging settings, we also include SpecBench~\cite{specbench}, HotpotQA~\cite{hotpotqa}, Polymath~\cite{polymath}, AIME24\&25~\cite{aime24}, and GPQA-Diamond~\cite{gpqa}. To prevent excessive GPU memory usage, we set the maximum generation length to 8192 tokens. Consistent with prior work~\cite{medusa,eagle3} on speculative decoding, we fix the batch size to 1. Detailed benchmark settings and analysis are in Appendix~\ref{sec:appendix-full-benchmark}. 

% Given the high inference cost of reasoning models, we randomly sample 100 examples from each benchmark to control overall evaluation time. For the Qwen3 model family, however, we additionally evaluate on the full set of benchmarks, with further details provided in Appendix~\ref{sec:appendix-full-benchmark}.

% \textbf{Metrics.}
% We report pass@1 accuracy for reasoning and LLM-judge scores for SpecBench. Efficiency is measured by throughput (tokens/s, TPS) and speedup relative to target-only decoding. Additionally, we define the target-invocation ratio (\(r_t\)) as the percentage of target model calls per generated token to quantify verification overhead.

\textbf{Metrics.}
We report pass@1 accuracy for tasks with verifiable answers and LLM-judge scores for SpecBench. Efficiency is measured by throughput (tokens/s) and speedup relative to target-only decoding. Additionally, we define the target-invocation ratio (\(r_t\)) as the percentage of target model calls per generated token to quantify reliance on the expensive target model.

\begin{table*}[t]
\centering
\caption{Accuracy and speedup of different methods on five reasoning benchmarks, evaluated with a sampling temperature of 1.}
% ``Target'' and ``Proxy'' denote the selected target and proxy verifiers, respectively.}
\scriptsize
\setlength{\tabcolsep}{2pt}  
\begin{tabular}{clcccccccccccc}
\toprule
& & \multicolumn{2}{c}{GSM8K} & \multicolumn{2}{c}{MATH500} & \multicolumn{2}{c}{GK23-en} & \multicolumn{2}{c}{HumanEval} & \multicolumn{2}{c}{MBPP} & \multicolumn{2}{c}{Avg} \\
\midrule
Model & \multicolumn{1}{c}{Method} & Acc. & Speedup & Acc. & Speedup & Acc. & Speedup & Acc. & Speedup & Acc. & Speedup & Acc. & Speedup \\ 
\midrule
\multirow{7}{*}{Qwen3} 
& Single Model (Target,32B)     & 98\% & 1.00$\times$ & 85\% & 1.00$\times$ & 76\% & 1.00$\times$ & 87\% & 1.00$\times$ & 76\% & 1.00$\times$ & 84.4\% & 1.00$\times$ \\ 
& Single Model (Proxy,1.7B)        & 88\% & 2.62$\times$ & 80\% & 2.84$\times$ & 73\% & 2.90$\times$ & 76\% & 2.90$\times$ & 61\% & 2.90$\times$ & 75.6\% {\tiny-8.8\%} & 2.83$\times$ \\
\cmidrule(lr){2-14}
& HASS     & 96\% & 2.61$\times$ & 86\% & 2.05$\times$ & 75\% & 2.05$\times$ & 88\% & 1.96$\times$ & 74\% & 1.94$\times$ & 83.8\% {\tiny-0.6\%} & 2.12$\times$\\
& \tscell{HASS + TriSpec} & \tscell{95\%} & \tscell{\textbf{3.01$\times$}} & \tscell{84\%} & \tscell{\textbf{2.51$\times$}} & \tscell{77\%} & \tscell{\textbf{2.54$\times$}} & \tscell{89\%} & \tscell{\textbf{2.31$\times$}} & \tscell{72\%} & \tscell{\textbf{2.25$\times$}} & \tscell{83.4\% {\tiny-1.0\%}} & \tscell{\textbf{2.52$\times$}}\\
\cmidrule(lr){2-14}
& EAGLE3     & 98\% & 2.78$\times$ & 84\% & 2.55$\times$ & 77\% & 2.52$\times$ & 86\% & 2.58$\times$ & 74\% & 2.52$\times$ & 83.8\% {\tiny-0.6\%} & 2.59$\times$ \\
& \tscell{EAGLE3 + TriSpec} & \tscell{96\%} & \tscell{\textbf{3.33$\times$}} & \tscell{86\%} & \tscell{\textbf{3.06$\times$}} & \tscell{76\%} & \tscell{\textbf{3.13$\times$}} & \tscell{88\%} & \tscell{\textbf{2.94$\times$}} & \tscell{73\%} & \tscell{\textbf{2.93$\times$}} & \tscell{83.8\% {\tiny-0.6\%}} & \tscell{\textbf{3.08$\times$}}\\
\midrule
\multirow{7}{*}{DSQ}
& Single Model (Target,32B)        & 92\% & 1.00$\times$ & 88\% & 1.00$\times$ & 77\% & 1.00$\times$ & 88\% & 1.00$\times$ & 76\% & 1.00$\times$ & 84.2\% & 1.00$\times$ \\
& Single Model (Proxy,1.5B)  & 71\% & 2.99$\times$ & 78\% & 2.91$\times$ & 66\% & 3.15$\times$ & 62\% & 3.12$\times$ & 47\% & 3.11$\times$ & 64.8\% {\tiny-19.4\%} & 3.06$\times$ \\
\cmidrule(lr){2-14}
& HASS      & 90\% & 2.97$\times$ & 87\% & 2.19$\times$ & 76\% & 2.33$\times$ & 90\% & 1.98$\times$ & 75\% & 2.04$\times$ & 83.6\% {\tiny-0.6\%} & 2.30$\times$ \\ 
& \tscell{HASS + TriSpec} & \tscell{89\%} & \tscell{\textbf{3.27$\times$}} & \tscell{87\%} & \tscell{\textbf{2.47$\times$}} & \tscell{77\%} & \tscell{\textbf{2.74$\times$}} & \tscell{87\%} & \tscell{\textbf{2.12$\times$}} & \tscell{73\%} & \tscell{\textbf{2.18$\times$}} & \tscell{82.6\% {\tiny-1.6\%}} & \tscell{\textbf{2.56$\times$}} \\
\cmidrule(lr){2-14}
& EAGLE3    & 92\% & 3.14$\times$ & 85\% & 2.40$\times$ & 77\% & 2.53$\times$ & 89\% & 2.23$\times$ & 76\% & 2.29$\times$ & 83.8\% {\tiny-0.4\%} & 2.52$\times$ \\
& \tscell{EAGLE3 + TriSpec} & \tscell{88\%} & \tscell{\textbf{3.40$\times$}} & \tscell{87\%} & \tscell{\textbf{2.79$\times$}} & \tscell{79\%} & \tscell{\textbf{2.92$\times$}} & \tscell{88\%} & \tscell{\textbf{2.36$\times$}} & \tscell{72\%} & \tscell{\textbf{2.42$\times$}} & \tscell{82.8\% {\tiny-1.4\%}} & \tscell{\textbf{2.78$\times$}} \\
\midrule
\multirow{5}{*}{DSL}
& Single Model (Target,70B) & 87\% & 1.00$\times$ & 88\% & 1.00$\times$ & 75\% & 1.00$\times$ & 87\% & 1.00$\times$ & 80\% & 1.00$\times$ & 83.4\% & 1.00$\times$ \\ 
& Single Model (Proxy,8B)  & 72\% & 3.80$\times$ & 81\% & 3.92$\times$ & 70\% & 3.95$\times$ & 79\% & 3.93$\times$ & 60\% & 3.96$\times$ & 72.4\% {\tiny-11.0\%} & 3.91$\times$ \\
\cmidrule(lr){2-14}
& EAGLE3    & 85\% & 2.79$\times$ & 88\% & 1.98$\times$ & 76\% & 2.05$\times$ & 88\% & 2.01$\times$ & 81\% & 1.96$\times$ & 83.2\% {\tiny-0.2\%} & 2.16$\times$ \\
& \tscell{EAGLE3 + TriSpec} & \tscell{86\%} & \tscell{\textbf{3.93$\times$}} & \tscell{89\%} & \tscell{\textbf{2.74$\times$}} & \tscell{78\%} & \tscell{\textbf{3.06$\times$}} & \tscell{89\%} & \tscell{\textbf{2.32$\times$}} & \tscell{78\%} & \tscell{\textbf{2.39$\times$}} & \tscell{84.0\% {\tiny+0.6\%}} & \tscell{\textbf{2.89$\times$}} \\
\bottomrule
\end{tabular}
\label{tab:temp1}
% \vspace{-10pt}
\end{table*}

% \rowcolor{cyan!10}

% \begin{table*}[ht]
% \color{blue}
% \centering
% \caption{\textcolor{blue}{Accuracy and speed in general-domain and extremely difficult scenarios.}}
% \footnotesize  
% \begin{tabular}{clcccccc}
% \toprule
% & & \multicolumn{2}{c}{SpecBench} & \multicolumn{2}{c}{Polymath} & \multicolumn{2}{c}{HotpotQA} \\
% \midrule
% & Method & Score & Speedup & Accuracy & Speedup & Accuracy & Speedup \\
% \midrule
% \multirow{5}{*}{Qwen3} 
% & Single Model (Target, 32B) & 8.82& 1.00$\times$ & 27.8\% & 1.00$\times$ & 93.0 & 1.00$\times$ \\
% & Single Model (Proxy, 1.7B) & 7.33& 2.72$\times$ & 9.4\% & 3.35$\times$ & 84.2 & 2.96$\times$ \\
% \cmidrule(lr){2-8}
% & EAGLE3 & 8.77& 2.80x& 27.20\% & 2.34$\times$ & 92.8 & 2.77x\\
% & EAGLE3+TriSpec & 8.62& 3.17x& 26.11\% & 3.11$\times$ & 92.8 & 3.58$\times$ \\
% \bottomrule
% \label{tab:new_benchmark}
% \end{tabular}
% \end{table*}

\begin{table*}[t!]
\centering
\caption{Performance and speedup on general-domain and more challenging reasoning benchmarks.}
% \vspace{-5pt}
% \footnotesize  
\scriptsize 
\setlength{\tabcolsep}{1.5pt} 
\begin{tabular}{clcccccccccccc}
\toprule
& & \multicolumn{2}{c}{SpecBench} & \multicolumn{2}{c}{HotpotQA} & \multicolumn{2}{c}{Polymath} & \multicolumn{2}{c}{AIME24} & \multicolumn{2}{c}{AIME25} & \multicolumn{2}{c}{GPQA-Diamond} \\
\cmidrule(lr){1-14}
& Method 
& Score & Speedup 
& Acc. & Speedup 
& Acc. & Speedup 
& Acc. & Speedup 
& Acc. & Speedup 
& Acc. & Speedup \\
\midrule
\multirow{4}{*}{Qwen3} 
& Single Model (Target, 32B) 
& 8.82 & 1.00$\times$ 
& 93.0\% & 1.00$\times$ 
& 27.8\% & 1.00$\times$ 
& 24/30 & 1.00$\times$ 
& 22/30 & 1.00$\times$ 
& 67.2\% & 1.00$\times$ \\
& Single Model (Proxy, 1.7B) 
& 7.33 & 2.72$\times$ 
& 84.2\% & 2.96$\times$ 
& 9.4\% & 3.35$\times$ 
& 16/30 & 3.60$\times$ 
& 12/30 & 3.54$\times$ 
& 44.9\% & 3.33$\times$ \\
\cmidrule(lr){2-14}
& EAGLE3 
& 8.77 & 2.80$\times$ 
& 92.8\% & 2.77$\times$ 
& 27.2\% & 2.34$\times$ 
& 22/30 & 2.52$\times$ 
& 21/30 & 2.31$\times$ 
& 68.2\% & 2.97$\times$ \\
& \tscell{EAGLE3+TriSpec}
& \tscell{8.62} & \tscell{\textbf{3.17$\times$}} 
& \tscell{92.8\%} & \tscell{\textbf{3.58$\times$}} 
& \tscell{26.1\%} & \tscell{\textbf{3.11$\times$}} 
& \tscell{23/30} & \tscell{\textbf{3.37$\times$}} 
& \tscell{21/30} & \tscell{\textbf{3.30$\times$}} 
& \tscell{66.2\%} & \tscell{\textbf{4.39$\times$}} \\
\bottomrule
\label{tab:new_benchmark}
\end{tabular}
\vspace{-12pt}
\end{table*}

\subsection{Effectiveness of TriSpec}
% Table~\ref{tab:model_performance} reports the performance and efficiency of TriSpec at a sampling temperature of 0. On all benchmark, TriSpec preserves predictive accuracy with an average degradation within 1\%. Concurrently, proxy-side verification delivers significantly higher throughput than standard SD. Specifically, for Qwen3, TriSpec enhances the speedup of HASS from 2.30$\times$ to 3.26$\times$ and EAGLE3 from 2.77$\times$ to 3.55$\times$, i.e., roughly a 30\% relative improvement. Similar improvements (15\%-35\%) are observed across the DSQ and DSL families.
% In contrast, while alternative lossy baselines like SpecCascade offer tunable trade-offs, they generally yield a less optimal balance between accuracy and speedup compared to TriSpec. Additional results under stochastic decoding with temperature $1.0$ are provided in Appendix~\ref{app:temp1_results}, where TriSpec remains robust and consistently improves throughput while maintaining competitive accuracy.
Table~\ref{tab:model_performance} reports the accuracy and efficiency of TriSpec at temperature 0. Across all benchmarks, TriSpec preserves accuracy with less than 1\% average degradation while substantially improving throughput over standard SD. For Qwen3, it increases the average speedup of HASS from 2.35$\times$ to 3.22$\times$ and EAGLE3 from 2.85$\times$ to 3.55$\times$, yielding relative gains of approximately 25\%--37\%, with consistent improvements observed on DSQ and DSL. Compared with lossy SpecCascade variants, TriSpec achieves a better accuracy--speed trade-off. Notably, on Qwen3, EAGLE3 + TriSpec achieves the same average accuracy of 84.2\% as EAGLE3 + SC[Token], while improving the average speedup from 2.96$\times$ to 3.55$\times$. These results demonstrate that TriSpec provides substantial additional acceleration with minimal task-level performance degradation.

Results at temperature $1.0$ (Table~\ref{tab:temp1}) further demonstrate the effectiveness of TriSpec under stochastic decoding. With EAGLE-3, TriSpec increases the average speedup from 2.59$\times$ to 3.08$\times$ on Qwen3, from 2.52$\times$ to 2.78$\times$ on DSQ, and from 2.16$\times$ to 2.89$\times$ on DSL, while maintaining average accuracy within 1.4 percentage points of target-only decoding across these model families.

% Table~\ref{tab:efficiency} provides a detailed breakdown of verification efficiency on MATH500 and HumanEval. Compared to standard SD, TriSpec consistently reduces the target-invocation ratio ($r_t$), per-round verification time ($t_v$), and end-to-end latency ($L$). For instance, on the Qwen3 family, TriSpec nearly halves $r_t$ while decreasing $t_v$ from 0.094s to 0.070s, resulting in significantly lower latency. These consistent improvements across all model families demonstrate TriSpec’s effectiveness in shifting the verification workload away from the expensive target model.

TriSpec remains effective beyond standard reasoning tasks, generalizing to broader domains and more challenging reasoning settings (Table~\ref{tab:new_benchmark}). On HotpotQA, TriSpec improves EAGLE-3 speedup from 2.77$\times$ to 3.58$\times$ while maintaining the same 92.8\% accuracy. It also achieves substantial gains on challenging benchmarks such as AIME24 and GPQA-Diamond, increasing speedup from 2.52$\times$ to 3.37$\times$ and from 2.97$\times$ to 4.39$\times$, respectively, with only minor accuracy variations. On SpecBench, TriSpec maintains competitive performance while improving speedup across subtasks, as shown in Figure~\ref{fig:combine_ablation}(a). Further details are provided in Appendix~\ref{sec:appendix-new-benchmark}.

Table~\ref{tab:efficiency} breaks down the verification efficiency of TriSpec on MATH500 and HumanEval. Compared with standard SD, TriSpec consistently reduces the target-invocation ratio ($r_t$), average per-round verification time ($t_v$), and end-to-end latency ($L$), while maintaining comparable acceptance lengths ($\tau$). For example, on Qwen3, it nearly halves $r_t$ and reduces $t_v$ from 0.094s to 0.070s. This indicates that the savings from fewer target invocations outweigh the additional proxy overhead, translating into end-to-end acceleration without relying on longer draft acceptance. Similar trends across DSQ and DSL further demonstrate the effectiveness of proxy verification across model families.

We further evaluate the additional memory overhead introduced by TriSpec's proxy verifier. In our Qwen3-series experiments, the proxy requires substantially less memory than the target model across all context lengths. It accounts for only about 6.4\% of the total memory at 8k context and remains below 15\% even at 128k, indicating that proxy verification introduces only modest additional memory cost. Detailed measurements are provided in Appendix~\ref{app:memory_usage}.

% \begin{table*}[ht]
% \centering
% \caption{Comparison of per-sample averages across all methods on MATH500 and HumanEval with the Qwen3 family: Target-invocation ratio ($T_r$), activated parameters per accepted token ($P_a$), Latency, and Speed).}
% \scriptsize
% \setlength{\tabcolsep}{4pt}  
% \begin{tabular}{lcccccccc}
% \toprule
% & \multicolumn{4}{c}{MATH500} & \multicolumn{4}{c}{HumanEval} \\
% \midrule
% \multicolumn{1}{c}{Method} & $T_r$ & $P_a$(B) & Latency(s) & Speed(Tok/s) & $T_r$ & $P_a$(B) & Latency(s) & Speed(Tok/s) \\
% \midrule

% Single Model (Tar.,32B) & 100.00\% & / & 258.76 & 14.24 & 100.00\% & / & 234.27 & 14.65 \\
% \midrule
% HASS & 25.01\% & 8.00 & 115.04 & 32.45 & 27.14\% & 8.68 & 119.32 & 29.80 \\
% HASS + confidence relax & 19.77\% & 6.32 & 177.92 & 37.06 & 24.78\% & 7.93 & 161.20 & 31.30 \\
% HASS + TriSpec & \textbf{10.24\%} & \textbf{3.68} &	\textbf{83.03} & \textbf{45.26} & \textbf{11.61\%} & \textbf{4.11} &	\textbf{78.74} & \textbf{43.38} \\
% \midrule
% EAGLE3 & 21.38\% & 6.84 & 97.52 & 38.96 & 22.05\% & 7.06 & 86.62 & 39.76 \\
% EAGLE3 + confidence relax & 19.58\% & 6.26 &	113.98 & 41.45 & 20.60\% & 6.59 & 103.48 & 41.35 \\
% EAGLE3 + TriSpec & \textbf{9.95\%} & \textbf{3.56} & \textbf{74.73} & \textbf{49.74} & \textbf{11.15\%} & \textbf{3.93} & \textbf{71.83} & \textbf{48.00} \\

% \bottomrule
% \end{tabular}
% \label{tab:efficiency}
% \end{table*}

\begin{table*}[t!]
\centering
\caption{Per-sample averages of methods on MATH500 and HumanEval: Target-invocation ratio ($r_t$), average per-round verification time ($t_v$), average acceptance length ($\tau$), Latency($L$), and Speed.}
% \footnotesize
\scriptsize
\setlength{\tabcolsep}{3.5pt}  
\begin{tabular}{clcccccccccc}
\toprule
& & \multicolumn{5}{c}{MATH500} & \multicolumn{5}{c}{HumanEval} \\
\midrule
Model & \multicolumn{1}{c}{Method} & $r_t$ & $t_v$(s) & $\tau$ & $L$(s) & Speed(Tok/s) & $r_t$ & $t_v$(s) & $\tau$ & $L$(s) & Speed(Tok/s) \\
\midrule
\multirow{6}{*}{Qwen3} 
& Single Model (Target,32B) & 100.00\% & / & / & 258.76 & 14.24 & 100.00\% & / & / & 234.27 & 14.65 \\
\cmidrule(lr){2-12}
& HASS & 25.01\% & 0.094& 4.21 & 115.04 & 32.45 & 27.14\% & 0.103 & 3.93 & 119.32 & 29.80 \\
& \tscell{HASS + TriSpec} & \tscell{\textbf{10.24\%}} & \tscell{\textbf{0.070}} & \tscell{\textbf{4.39}} & \tscell{\textbf{83.03}} & \tscell{\textbf{45.26}} & \tscell{\textbf{11.61\%}} & \tscell{\textbf{0.077}} & \tscell{\textbf{4.41}} & \tscell{\textbf{78.74}} & \tscell{\textbf{43.38}} \\
\cmidrule(lr){2-12}
& EAGLE3 & 21.38\% & 0.094 & \textbf{4.72} & 97.52 & 38.96 & 22.05\% & 0.103 & 4.60 & 86.62 & 39.76 \\
& \tscell{EAGLE3 + TriSpec} & \tscell{\textbf{9.95\%}} & \tscell{\textbf{0.071}} & \tscell{4.59} & \tscell{\textbf{74.73}} & \tscell{\textbf{49.74}} & \tscell{\textbf{11.15\%}} & \tscell{\textbf{0.080}} & \tscell{\textbf{4.76}} & \tscell{\textbf{71.83}} & \tscell{\textbf{48.00}} \\
\midrule
\multirow{6}{*}{DSQ} 
& Single Model (Target,32B) & 100.00\% & / & / & 156.82 & 16.82 & 100.00\% & / & / & 163.68 & 17.35 \\
\cmidrule(lr){2-12}
& HASS & 22.59\% & 0.068 & \textbf{4.50} & 47.92 & 50.29 & 23.55\% & 0.072& \textbf{4.16} & 56.75 & 48.40 \\
& \tscell{HASS + TriSpec} & \tscell{\textbf{11.10\%}} & \tscell{\textbf{0.054}} & \tscell{4.31} & \tscell{\textbf{43.25}} & \tscell{\textbf{57.30}} & \tscell{\textbf{14.20\%}} & \tscell{\textbf{0.061}} & \tscell{4.05} & \tscell{\textbf{52.90}} & \tscell{\textbf{51.15}} \\
\cmidrule(lr){2-12}
& EAGLE3 & 20.67\% & 0.068 & \textbf{4.79} & 46.13 & 53.77 & 21.45\% & 0.072& \textbf{4.54} & 51.80 & 53.66 \\
& \tscell{EAGLE3 + TriSpec} & \tscell{\textbf{10.63\%}} & \tscell{\textbf{0.055}} & \tscell{4.78} & \tscell{\textbf{42.06}} & \tscell{\textbf{60.03}} & \tscell{\textbf{13.25\%}} & \tscell{\textbf{0.064}} & \tscell{\textbf{4.54}} & \tscell{\textbf{48.92}} & \tscell{\textbf{57.02}} \\
\midrule
\multirow{4}{*}{DSL} 
& Single Model (Target,70B) & 100.00\% & / & / & 256.79 & 8.22 & 100.00\% & / & / & 254.18 & 8.37 \\
\cmidrule(lr){2-12}
& EAGLE3 & 28.70\% & 0.144 & 3.48 & 108.89 & 19.33 & 26.88\% & 0.147 & \textbf{3.66} & 107.92 & 20.84 \\
& \tscell{EAGLE3 + TriSpec} & \tscell{\textbf{11.60\%}} & \tscell{\textbf{0.093}} & \tscell{\textbf{3.64}} & \tscell{\textbf{78.32}} & \tscell{\textbf{27.54}} & \tscell{\textbf{15.18\%}} & \tscell{\textbf{0.114}} & \tscell{\textbf{3.59}} & \tscell{\textbf{87.51}} & \tscell{\textbf{25.70}} \\
\bottomrule
\end{tabular}
\vspace{-10pt}
\label{tab:efficiency}
\end{table*}

\subsection{Routing Strategy}
\begin{figure*}[t]
    \centering
    % \vspace{-5pt}
    \includegraphics[width=1\linewidth]{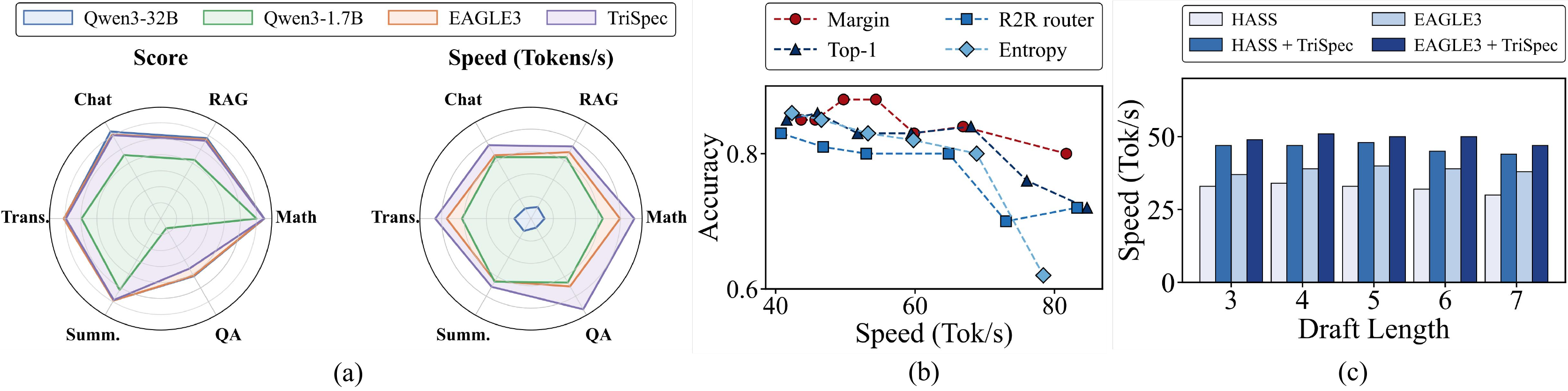}
    \vspace{-15pt}
    \caption{(a) Radar chart illustrating score and speed across all SpecBench subtasks. All experiments are conducted using models from the Qwen3 family. (b) Accuracy–speed trade-offs of different routing strategies. (c) Decoding speed under varying draft lengths for standard speculative decoding and TriSpec. All experiments are conducted on the MATH500 and using Qwen3-32B as the target.}
    \vspace{-10pt}
    \label{fig:combine_ablation}
\end{figure*}

% We examine both rule-based and training-based routing approaches. To avoid semantic bias from limited-capacity draft models, all routing designs rely on features from the proxy verifier rather than the draft model. Specifically, we evaluate top-1 probability, a composite entropy criterion~\cite{aasd}, and the probability margin between the top-two candidates. For training-based routing, we follow the R2R~\cite{r2r} to train a lightweight router for cross-model coordination. As shown in Figure~\ref{fig:combine_ablation}(b), while all methods can preserve accuracy under calibrated thresholds, margin-based routing offers the most favorable accuracy--speedup trade-off and is thus adopted as the default for TriSpec.
\textbf{Criterion Selection.}
We compare rule-based and training-based routing criteria, all using proxy-verifier features to avoid potential semantic bias from the limited-capacity drafter. Rule-based criteria include top-1 probability, composite entropy~\cite{aasd}, and the top-1--top-2 probability margin, while the training-based baseline follows R2R~\cite{r2r} with a lightweight router for cross-model coordination. As shown in Figure~\ref{fig:combine_ablation}(b), although all criteria can preserve accuracy after threshold calibration, margin-based routing achieves the best accuracy--speed trade-off and is used as TriSpec's default.

\begin{figure}[t]
    \centering
    \includegraphics[width=1\linewidth]{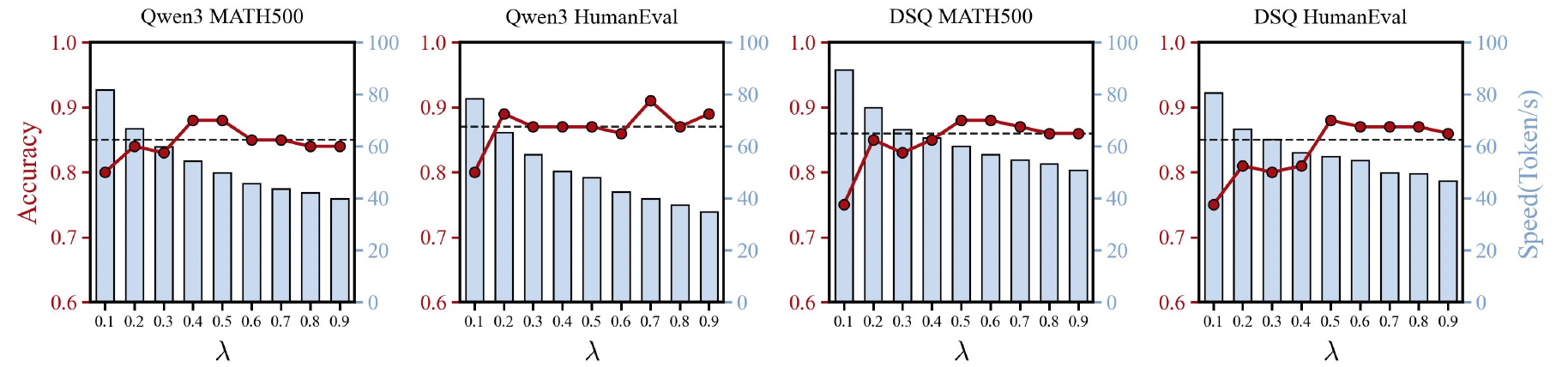}
    \vspace{-10pt}
    \caption{Accuracy–speed trade-off of TriSpec under different margin thresholds \( \lambda \), where the dashed line denotes the target-only baseline accuracy. }
    \vspace{-12pt}
    \label{fig:ablation_lambda}
\end{figure}

\textbf{Threshold Sensitivity.}
We further examine the impact of the margin threshold $\lambda$, which controls when TriSpec escalates proxy verification to the target model. As shown in Figure~\ref{fig:ablation_lambda}, larger $\lambda$ values induce more conservative routing, gradually reducing speed while stabilizing accuracy by filtering low-confidence proxy decisions. This consistent trend across models and tasks shows that $\lambda$ provides an effective control knob for navigating the efficiency--accuracy trade-off. Empirically, $\lambda=0.5$ is a robust general default, while task-specific calibration can further improve the trade-off.

% \begin{table*}[t]
% \centering
% \caption{Performance under different draft model training strategies on the Qwen3 family. TP denotes Training Parameters in billions. }
% \small
% % \setlength{\tabcolsep}{4pt}  
% \begin{tabular}{lcccccccc}
% \toprule
% & & & \multicolumn{3}{c}{MATH500} & \multicolumn{3}{c}{HumanEval} \\
% \midrule
% \multicolumn{1}{c}{Method} & Training Strategy & TP & $\tau$ & $r_t$ & Speedup &  $\tau$ & $r_t$ & Speedup \\
% \midrule

% HASS & / & 0.21 & 4.21 & 25.01\% & 2.27$\times$ & 3.93 & 27.14\% & 2.03$\times$ \\
% HASS + TriSpec & joint training from scratch & 0.26 & 4.39 & 10.24\% & 3.18$\times$ & 4.42 & 11.61\% & 2.99$\times$ \\
% HASS + TriSpec & adapter-only finetuning & 0.05 & 4.31 & 10.38\% & 3.08$\times$ & 4.37 &	11.76\% & 2.95$\times$ \\
% \midrule
% EAGLE3 & / & 1.35 & 4.72 & 21.38\% & 2.73$\times$ & 4.60 & 22.05\% & 2.71$\times$ \\
% EAGLE3 + TriSpec & joint training from scratch & 1.77 & 4.59 & 9.95\% & 3.49$\times$ & 4.76 & 11.15\% & 3.27$\times$ \\
% EAGLE3 + TriSpec & adapter-only finetuning & 0.42 & 4.52 & 9.99\% & 3.44$\times$ & 4.65 & 11.40\% & 3.21$\times$ \\

% \bottomrule
% \end{tabular}
% \vspace{-1em}
% \label{tab:adapter_training}
% \end{table*}

\begin{table*}[t]
% \vspace{-10pt}
\centering
% \caption{Performance under different draft model training strategies on the Qwen3 family. TP denotes Training Parameters in billions.}
\caption{Ablations on the Qwen3 family. Left: comparing draft-model training strategies, with TP denoting trainable parameters in billions. Right: average performance with/without token pruning.}
\vspace{-5pt}
\label{tab:ablation}
\footnotesize
\setlength{\tabcolsep}{2pt}
\begin{tabular}{lccccccc}
\toprule
& & \multicolumn{3}{c}{MATH500} & \multicolumn{3}{c}{HumanEval} \\
\cmidrule(lr){3-5} \cmidrule(lr){6-8}
Training Strategy & TP & $\tau$ & $r_t$ & Speedup & $\tau$ & $r_t$ & Speedup \\
\midrule
joint from scratch & 1.77 & 4.59 & 9.95\% & 3.49$\times$ & 4.76 & 11.1\% & 3.27$\times$ \\
adapter finetuning   & 0.42 & 4.52 & 9.99\% & 3.44$\times$ & 4.65 & 11.4\% & 3.21$\times$ \\
\bottomrule
\end{tabular}
\hspace{6pt}
\begin{tabular}{cccc}
\toprule
& \multicolumn{3}{c}{five datasets (avg)} \\
\cmidrule(lr){2-4}
Verification strategy & $\tau$ & Acc. & Speedup \\
\midrule
w/ Token pruning  & 4.71 & 84.0\% & 3.55$\times$ \\
w/o Token pruning & 4.58 & 84.2\% & 3.46$\times$ \\
\bottomrule
\end{tabular}
\vspace{-10pt}
\end{table*}

% \subsection{Draft model training strategies}
% \label{sec:draft-training}
% EAGLE-family draft models are parameter-efficient and relatively inexpensive to train. We recommend joint from-scratch training of the draft model backbone and the adapter, which best accommodates feature inputs from both the target and the proxy. When backbone weights are already available, we also consider adapter-only finetuning for TriSpec to reduce the training cost.

% To compare different strategies, we use acceptance length to measure draft quality and training parameters (TP) to assess the training cost.
% As shown in Table~\ref{tab:adapter_training}, the two strategies yield very similar outcomes: joint from-scratch training attains slightly longer acceptance lengths, while adapter-only finetuning achieves comparable quality with lower training cost. In all cases, the resulting systems deliver end-to-end speed that substantially exceeds standard SD verification.

\subsection{Ablation Studies}
\label{sec:draft-training}
% The ablation studies focus on two key aspects of TriSpec: (i) the tree prune strategy and (ii) the draft length per round.

% \textbf{Margin threshold.}
% The margin threshold \(\lambda\) controls the accuracy–speed trade-off by deciding when the proxy hands verification to the target. 
% % We instantiate routing with the assistant’s probability margin between the top-1 and top-2 candidates and vary the threshold \(\lambda\). 
% Larger \(\lambda\) imposes stricter routing and increases target invocations; smaller \(\lambda\) routes more cases to the proxy and improves speed. Figure~\ref{fig:ablation_route} reports accuracy and speedup across \(\lambda\). As \(\lambda\) decreases, throughput increases; once \(\lambda\) approaches \(0.5\), accuracy begins to degrade and gradually approaches the proxy-only baseline.

% \textbf{Draft Training Strategies.} 
% TriSpec supports two training paradigms for EAGLE-family draft models to balance draft quality and computational overhead. Joint from-scratch training of the backbone and adapter captures the synergy between target and proxy features for optimal performance, while adapter-only finetuning provides a cost-effective alternative when pre-trained weights are available. As shown in Table~\ref{tab:ablation} left, while joint training offers slightly longer acceptance lengths, both strategies consistently deliver end-to-end speeds that substantially exceed standard speculative decoding.

\textbf{Draft length.}
% The per-round draft length affects both verification routing and attainable speedup. Due to proxy-target behavioral mismatches and token pruning, TriSpec's optimal draft length can differ from standard SD. Figure~\ref{fig:combine_ablation}(c) compares throughput for both methods across draft lengths. For Qwen3 on \textsc{MATH500}, TriSpec is faster at all lengths, while its optimal setting shifts modestly relative to standard SD: from 4 to 5 for HASS-trained drafters and from 5 to 4 for EAGLE3.
Per-round draft length affects both routing decisions and attainable speedup. Since proxy-target mismatches and token pruning change the verification dynamics, TriSpec's optimal length may differ from standard SD. Figure~\ref{fig:combine_ablation}(c) compares throughput across draft lengths using Qwen3 on the \textsc{MATH500} benchmark. TriSpec outperforms standard SD at all lengths, with a modest shift in the optimum: from 4 to 5 for HASS-trained drafters and from 5 to 4 for EAGLE3.

\textbf{Draft Training Strategies.}
TriSpec supports two training paradigms for EAGLE-family drafters to balance performance and training cost. Joint training from scratch optimizes the drafter backbone and adapter together, using 1.77B trainable parameters, while adapter finetuning freezes the pretrained drafter and updates only the 0.42B-parameter adapter. As shown in the left part of Table~\ref{tab:ablation}, joint training yields slightly longer acceptance lengths and higher speedups, but adapter finetuning remains close, offering a cost-effective alternative when pretrained weights are available.

% \textbf{Token pruning strategies.}
% Under TriSpec's Case II, the target model is involved in the verification process. Leveraging the parallelized nature of verification, the target can either validate the entire draft or trust the proxy’s partial verification and only process the remaining tokens, a technique we refer to as token pruning. As demonstrated in Table~\ref{tab:ablation} right, the token pruning strategy enhances the average acceptance length without compromising predictive accuracy. This further underscores the reliability of proxy-side assessments when the proxy exhibits high confidence, allowing the target model to safely bypass redundant computations.

\textbf{Token Pruning Strategies.}
In TriSpec's Case II, the target model is invoked for verification. It can either re-verify the entire draft or trust the proxy-verified prefix and process only remaining tokens, which we call token pruning. As shown in the right part of Table~\ref{tab:ablation}, token pruning reduces target-side computation and extends the accepted prefix, improving speedup with only a slight accuracy drop. The drop is consistent with Section~\ref{sec:error_analysis}: pruning increases proxy-finalized decisions and may induce a small task-level gap. Thus, it provides an optional efficiency--accuracy trade-off for deployment.

\section{Related Work}
Beyond the approaches discussed earlier, several lines of work accelerate SD in specific settings. On the drafting side, retrieval-based methods like REST~\cite{rest} and LLMA~\cite{llma} match relevant input segments to propose draft tokens; cache-reuse methods, including GLIDE~\cite{glide} and MoA~\cite{moa}, exploit target KV caches for fast drafting; and layer-skipping methods~\cite{layerskip,draftandverify} build simplified draft models by partially reusing target parameters. AdaEAGLE~\cite{adaeagle} and SpecDec++~\cite{specdec++} further dynamically adjust draft length to reduce unnecessary computation. On the verification side, Judge Decoding~\cite{judgedecoding} trains a small judge to accept correct but misaligned drafts, while Cactus~\cite{hao2026cactus} improves draft acceptance by adjusting the verifier distribution. SPRINTER~\cite{sprinter} uses a lightweight classifier for sequential approximate verification, whereas VIA-SD~\cite{xian2026via} employs a target-derived slim verifier for hierarchical verification. For long-context acceleration, SpecPV~\cite{tan2026specpv} uses partial KV states for efficient verification, while TriForce~\cite{triforce} combines retrieval-based drafting with hierarchical speculation. Despite these advances, most methods relax acceptance or approximate verification, without fully exploiting the inherent alignment between same-family models to facilitate verification.

Small-large model collaboration has also been studied for acceleration. Query-level routing methods~\cite{routellm,graphrouter} select an SLM or LLM per request, R2R~\cite{r2r} learns token-level routing to correct divergent SLM tokens, and SpecCascade~\cite{speculativecascade} combines cascaded deferral with speculative execution. In contrast, TriSpec extends this principle to speculative decoding, where an efficient SLM proxy coordinates with the target and drafter for fine-grained verification and fewer target invocations.

\section{Conclusion}
In this work, we introduced a new perspective for accelerating speculative decoding by reducing verification time and proposed TriSpec, a ternary framework that combines a lightweight proxy verifier with the drafter and target model. By leveraging same-family smaller models as proxies, TriSpec exploits their strong alignment and reliable outputs to offload a substantial fraction of verification from the target, while a margin-based routing mechanism ensures accuracy is preserved through selective escalation. Extensive experiments show that TriSpec achieves up to 35\% additional speedup over standard speculative decoding and sustains accuracy with minimal loss. These results highlight the potential of proxy verification as a new and effective direction for speculative decoding.

\begin{ack}
HYJ and JCY are supported by National Natural Science Foundation of China
(No. 62676245), the Lingang Laboratory (Grant No. LGL-8889-01), STCSM
(No. 22DZ2229005), and the 111 Plan (No. BP0719010).
XFC is supported by National Natural Science Foundation of China (No. 62476109).
BH was supported by RGC Research Fellow Scheme No. RFS2627-2S02,
RGC General Research Fund No. 12202026,
RGC Young Collaborative Research Grant No. C2005-24Y,
RGC General Research Fund No. 12200725,
NSFC Major Research Plan No. 92570109,
NSFC General Program No. 62376235,
Guangdong and Hong Kong Universities ``1+1+1'' Joint Research Collaboration
Scheme No. JRCS-25-26-COMP-05,
HKBU Faculty Niche Research Areas No. RC-FNRA-IG/25-26/SCI/05,
and HKBU CSD Departmental Incentive Scheme.

\end{ack}

{
    \bibliography{TriSpec}
    \bibliographystyle{plainnat}
}

\newpage
\appendix
\onecolumn
% \section{LLM Usage Statement}
% Below, we clarify how Large Language Models (LLMs) are used in this study. During manuscript preparation, LLMs served only as broad writing assistants; they did not contribute to idea formation or to drafting the technical substance. In our methodology, LLMs form a central component. Specifically, our approach is designed to accelerate LLM usage; we rely on LLMs (e.g., Qwen3-32B) both for constructing training data and for generating answers used in evaluation. Further specifics appear in the "Experiments" section of the main paper. The authors take full responsibility for all content written under their name.

\section{Algorithm}
Algorithm~\ref{alg:trispec} presents the formal pseudo-code for a single generation round of TriSpec, detailing the autoregressive drafting process, proxy-based pre-verification, and margin-based routing mechanism for conditional escalation to the target model.

\begin{algorithm}[H]
\caption{TriSpec in a single round}
\label{alg:trispec}
\begin{algorithmic}[1]
\STATE \textbf{Inputs:} Draft model $\mathcal{M}_d$, Proxy verifier $\mathcal{M}_p$, Target model $\mathcal{M}_t$, input prefix $X$, margin-based threshold $\lambda$

\LineComment{Sample k draft tokens from $\mathcal{M}_d$ autoregressively}

\FOR{$i = 1$ to $k$} 
    \STATE $\mathbf{p}^{(d)}_i \gets \mathcal{M}_d(X + [x_1, \dots, x_{i-1}])$ 
    \STATE $x_i \sim \mathbf{p}^{(d)}_i$ 
\ENDFOR

\LineComment{Pre-verification by $\mathcal{M}_p$}

\STATE $\tau_{a}, \; [\mathbf{p}^{(p)}_1, \dots, \mathbf{p}^{(p)}_{k+1}] \gets \text{Verify}(\mathcal{M}_p, X+[x_1, \dots, x_{k}])$ 

\LineComment{Determine proxy-trusted length via margin-based rule}

\STATE $g(\mathbf{p}^{(p)}_i) 
\gets \mathbf{1}\!\left\{ \operatorname{top}_1(\mathbf{p}^{(p)}_i) 
- \operatorname{top}_2(\mathbf{p}^{(p)}_i) \ge \lambda \right\},\; i=1,\ldots,k+1$ 
\STATE $\tau_m \gets \min(\{\, i \;|\; 0 \le i \le k,\; g(\mathbf{p}^{(p)}_{i+1})=0 \} \cup \{k+1\})$ 

\LineComment{Conditional routing to target based on proxy reliability}

\IF{$\tau_a < \tau_m$} 
    \STATE $x_{\mathrm{bonus}} \sim \mathbf{p}^{(p)}_{\tau_a+1} $ 
    \STATE \textbf{return} $X + [x_1, \dots, x_{\tau_a},x_{\mathrm{bonus}}]$
\ELSE 
    \STATE $\tau_{t}, \; [\mathbf{p}^{(t)}_{\tau_m+1}, \dots, \mathbf{p}^{(t)}_{k+1}] \gets \text{Verify}(\mathcal{M}_t, X+[x_1, \dots, x_{k}])$ 
    \STATE $x_{\mathrm{bonus}} \sim \mathbf{p}^{(t)}_{\tau_m+\tau_t+1} $ 
    \STATE \textbf{return} $X + [x_1, \dots, x_{\tau_m+\tau_t},x_{\mathrm{bonus}}]$
\ENDIF
\end{algorithmic}
\end{algorithm}

\section{Experimental Details and Additional Experimental Results}
\label{sec:Experimental Details}

% \subsection{Details of token-level alignment experiments}
% \label{sec:appendix-token-level-experiments}
% This section provides the experimental details for Figure~\ref{fig:latency_breakdown}, which evaluates token-level alignment between the proxy verifier and the target model. We use ShareGPT as the reference dataset and first generate autoregressive responses using Qwen3-32B as the target model. We then prefill the target completions with Qwen3-1.7B, the proxy verifier, and record the proxy's next-token prediction at each position. A token is counted as an exact match when the proxy and target predictions are identical. For non-matching positions, we further determine whether the proxy token can be categorized as a benign divergence. Specifically, we first check whether the proxy token lies within the target model's top-\(k\) candidates or top-\(p\) nucleus at the same position, with \(k=3\) and \(p=0.9\). If not, we consult Qwen3-235B as a stronger judge using the same top-\(k\) and top-\(p\) settings, mark the proxy token as benign if it appears in the resulting candidate set. All remaining mismatches are labeled as salient divergences. For both the proxy and target models, we use temperature \(=0\) to ensure deterministic outputs.

\subsection{Draft Model Training and Inference Details for TriSpec}
\label{sec:appedix-draft-training}
We train the draft models following the official HASS/EAGLE-3 codebases and the guidelines in their respective papers. In TriSpec, the drafter backbone and adapter are jointly trained from scratch by default. Specifically, HASS is trained on 68K ShareGPT samples, while EAGLE-3 additionally incorporates UltraChat-200K~\cite{ultrachat}. We first generate responses autoregressively using the target model, and then compute token-level hidden features offline from both the proxy and target models to supervise baseline drafter training and TriSpec adapter learning.

Training is conducted on 8$\times$ NVIDIA A100 80GB GPUs, using the official default hyperparameters unless otherwise specified. During EAGLE-3 training, we retain the full vocabulary and set the batch size to 1 to avoid out-of-memory errors. For a 32B target model, HASS/EAGLE-3 training typically takes 3--4 days, whereas TriSpec requires approximately 4--5 days due to the additional proxy-feature extraction and adapter training. Detailed training costs are reported in Table~\ref{training_cost}.

During inference, we follow the official EAGLE implementation scripts. All benchmark inference experiments are conducted on a single NVIDIA A100 80GB GPU. For KV-cache management, TriSpec retains the target cache across decoding rounds and, when target verification resumes, processes any uncached accepted tokens together with the current drafts in a single forward pass. This avoids recomputing the cached prefix, and the cache-update cost is included in the reported $t_v$.

\begin{table*}[ht]
\centering
\caption{Training cost of EAGLE3 and TriSpec on Qwen3 family}
\footnotesize  
\begin{tabular}{lccccccc}
\toprule
\multicolumn{1}{c}{Method} & Training Strategy & Data Size & Seq Len  & TP(B) &  Epochs & Training Time(h)  \\
\midrule
EAGLE3 & /  & 518K & 2048 & 1.35 &  6 & 78 \\
EAGLE3 + TriSpec & joint-training from scratch & 518K & 2048 & 1.77 & 6 & 115 \\
EAGLE3 + TriSpec & adapter-only finetune & 518K & 2048 & 0.42 & 3 & 44 \\
\bottomrule
\label{training_cost}
\end{tabular}
\end{table*}

\subsection{Baseline Implementation Details: SpecCascade}
\label{sec:appendix-baseline-detail}

We adopt \textsc{SpecCascade} as a main baseline, which introduces several rule-based strategies for relaxed verification in speculative decoding. In our experiments, we implement three representative \emph{deferral rules}:
\begin{itemize}
    \item \textbf{Chow rule}: This rule accepts a candidate token when the draft model assigns it a probability above a fixed threshold (see Table~1 in the original \textsc{SpecCascade} paper). Our \emph{confidence-filter} method in Appendix~\ref{sec:appendix-token-level-experiments} is equivalent to this rule.
    \item \textbf{Optimal deferral rule}: This rule compares the draft-model probability of the candidate token with the maximum probability under the target-model distribution. Its rejection threshold is determined by the distance between the draft and target distributions (see Eq.~(10) in the original paper).
    \item \textbf{Token-specific deferral rule}: This rule accepts a candidate token by comparing its probability under the target-model distribution with the target model's maximum probability, without using the draft-model distribution (see Eq.~(14) in the original paper).
\end{itemize}

We set $\alpha=0.05$ for the Chow and Optimal deferral rules, both of which rely on the draft-model distribution, and use a larger value of $\alpha=0.2$ for the token-specific deferral rule. As shown in Table~\ref{tab:model_performance}, when the drafter is a \emph{small single-layer} model, deferral rules based on draft-model probabilities substantially degrade accuracy: even with a small $\alpha$, Chow and Optimal deferral still incur notable performance drops, consistent with our token-level analysis in Appendix~\ref{sec:appendix-token-level-experiments}. By contrast, the token-specific deferral rule better preserves accuracy, since its acceptance decision is governed by the target-model distribution and therefore favors tokens that are more consistent with the target model's preferences. However, because a small singlelayer drafter can generate only a limited number of high-quality candidates, the additional speedup obtained by this rule remains modest.

% \subsection{Experiments on different temperature settings}
% \label{sec:appendix-new-temperature}
% To further assess the robustness of TriSpec under varying sampling temperatures, we conduct an additional evaluation with the sampling temperature set to 1.0. The corresponding results are summarized in Table~\ref{tab:temp1}. Even under this non-deterministic setting, TriSpec attains substantial speedups while incurring negligible degradation in accuracy.
% \input{Tables/table_temperature1}

\subsection{Memory Usage}
\label{app:memory_usage}

We measure the GPU memory usage of the target model, proxy verifier, and drafter on the Qwen3-series models with batch size 1. As shown in Table~\ref{tab:memory_usage}, the additional memory overhead introduced by the proxy verifier is small relative to the overall footprint. At an 8k context length, the proxy accounts for only 6.4\% of total memory, and the complete system still fits within a single A100 GPU. Even at a 128k context length, the proxy accounts for only 14.8\% of total memory, a setting where the acceleration benefit of speculative decoding is already substantially constrained by long-context verification and attention overhead. These results indicate that the proxy verifier introduces only modest memory overhead in practical settings.

\begin{table}[h]
\centering
\vspace{-5pt}
\caption{GPU memory usage (GB) of Qwen3-series models under different context lengths.}
\label{tab:memory_usage}
\vspace{2mm}
\footnotesize
\begin{tabular}{lccccccc}
\toprule
\textbf{Component} & \textbf{2k} & \textbf{4k} & \textbf{8k} & \textbf{16k} & \textbf{32k} & \textbf{64k} & \textbf{128k} \\
\midrule
Target  & 66.01 & 66.50 & 67.47 & 69.42 & 73.32 & 81.12 & 96.72 \\
Proxy   & 4.27  & 4.49  & 4.91  & 5.76  & 7.46  & 10.86 & 17.66 \\
Drafter & 4.16  & 4.16  & 4.18  & 4.21  & 4.28  & 4.41  & 4.67  \\
\midrule
Total & 74.44 & 75.15 & 76.56 & 79.39 & 85.06 & 96.39 & 119.05 \\
Proxy / Total & 5.7\% & 6.0\% & 6.4\% & 7.3\% & 8.8\% & 11.3\% & 14.8\% \\
\bottomrule
\end{tabular}
\end{table}

\subsection{Standard Reasoning Benchmark Details and Statistical Robustness Analysis}
\label{sec:appendix-full-benchmark}
% To assess the stability and reliability of our evaluation, we conduct two complementary experiments. First, we extend the evaluation from 100-sample subsets to the full benchmark datasets. Table~\ref{tab:full_benchmark} reports performance and efficiency results on the Qwen3 family models across the complete test sets. Second, we perform repeated evaluation on the \textsc{GSM8K} benchmark by drawing eight distinct random subsets of 100 samples and computing the mean results with error bars, as shown in Figure~\ref{fig:error_bars}. Error bars in all figures denote one standard deviation across the independent runs.

% Both the full-benchmark evaluation and the repeated-subset experiments consistently show that TriSpec achieves substantially higher generation speed compared to standard speculative decoding (SD), while incurring only minor accuracy degradation. These findings align with our main conclusions and indicate that the observed improvements are not an artifact of subset selection. Moreover, the narrow error bars in Figure~\ref{fig:error_bars} indicate that performance variations introduced by random subset sampling are minor, further supporting the robustness of the observed trends.
Due to the substantial inference cost of large-scale reasoning models, we primarily evaluate on randomly sampled subsets of 100 examples from each reasoning benchmark. These subsets are used for all experimental results reported in the main text, enabling broad and computationally feasible comparisons across multiple model families and diverse tasks. For the Qwen3 family, we further conduct evaluations on the full benchmark datasets to provide a more comprehensive performance profile. The complete-test-set results are reported in Table~\ref{tab:full_benchmark}, showing that the performance and efficiency trends observed on the 100-sample subsets remain consistent at full scale.

\begin{figure}
    \centering
    \includegraphics[width=\linewidth]{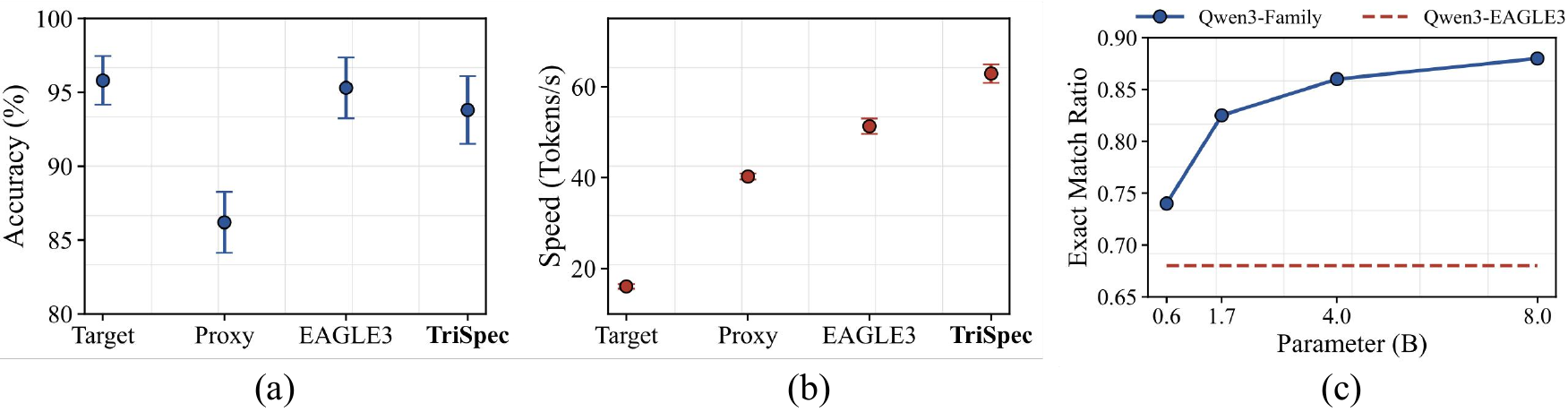}
    \caption{(a--b) Error bars indicate one standard deviation of accuracy and speed across repeated random subset evaluations for four generation methods. (c) Token-level exact-match ratios of Qwen3-family models at different parameter scales and the EAGLE-3 drafter, using Qwen3-32B predictions as the reference.}
    \label{fig:error_bars}
\end{figure}

We further assess the statistical robustness of the subset-based evaluation on GSM8K by drawing eight independent random subsets of 100 examples. As shown in Figure~\ref{fig:error_bars}(a--b), we report the mean accuracy and speed, with error bars indicating one standard deviation across the repeated subset samples. The narrow error bars show that the variation introduced by random sampling is small. Together with the full-benchmark results in Table~\ref{tab:full_benchmark}, these findings confirm that TriSpec consistently improves generation speed over standard speculative decoding while incurring only minor accuracy degradation, and that our conclusions are not an artifact of a particular subset selection.
\begin{table*}[ht]
\centering
\captionsetup[table]{skip=5pt}
\setlength{\tabcolsep}{2.5pt} 
\caption{Accuracy and speedup results on the full benchmark suite for Qwen3 family}
% \scriptsize  
\footnotesize
\begin{tabular}{lcccccccccc}
\toprule
& \multicolumn{2}{c}{GSM8K} & \multicolumn{2}{c}{MATH500} & \multicolumn{2}{c}{Gaokao2023en} & \multicolumn{2}{c}{Humaneval} & \multicolumn{2}{c}{MBPP}\\
\midrule
\multicolumn{1}{c}{Method} & Acc. & Speedup & Acc. & Speedup & Acc. & Speedup & Acc. & Speedup & Acc. & Speedup \\
\midrule
Single Model (Target, 32B) & 94.9\% & 1.00$\times$ & 86.8\% & 1.00$\times$ & 78.2\% & 1.00$\times$ & 86.6\% & 1.00$\times$ & 74.8\% & 1.00$\times$ \\
Single Model (Proxy, 1.7B) & 86.0\% & 2.52$\times$ & 78.2\% & 2.87$\times$ & 69.9\% & 2.76$\times$ & 77.4\% & 2.87$\times$ & 52.8\% & 2.93$\times$ \\
EAGLE3 & 94.3\% & 3.29$\times$ & 88.0\% & 2.77$\times$ & 77.1\% & 2.77$\times$ & 87.2\% & 2.72$\times$ & 74.2\% & 2.71$\times$ \\
EAGLE3 + TriSpec & 92.9\% & \textbf{4.08$\times$} & 87.4\% & \textbf{3.48$\times$} & 77.4\% & \textbf{3.39$\times$} & 86.0\% & \textbf{3.23$\times$} & 71.4\% & \textbf{3.33$\times$} \\
\bottomrule
\label{tab:full_benchmark}
\end{tabular}
\end{table*}

\subsection{Implementation Details for General-Domain and Challenging Reasoning Benchmarks}
\label{sec:appendix-new-benchmark}

This section describes the benchmarks and implementation protocols used to evaluate TriSpec on general-domain tasks and frontier-level reasoning challenges, including SpecBench, HotpotQA, Polymath, AIME 2024, AIME 2025, and GPQA-Diamond.

SpecBench contains 480 samples across six categories: multi-turn conversation, translation, summarization, question answering (QA), mathematical reasoning, and retrieval-augmented generation (RAG). Its multi-turn conversation category further covers diverse open-ended tasks such as writing, roleplay, extraction, and humanities, providing a broad evaluation of model performance across domains. HotpotQA is a natural-language multi-hop QA benchmark with explicit supporting-fact supervision, offering a QA format distinct from SpecBench; we evaluate on a randomly selected subset of 500 samples. Polymath is a multilingual mathematical reasoning benchmark covering 18 languages and four difficulty levels. To evaluate performance in multilingual and highly challenging settings, we sample 20 high-difficulty instances from each language, resulting in 360 test samples.

AIME 2024 and AIME 2025 are challenging mathematical reasoning benchmarks drawn from the American Invitational Mathematics Examination, each consisting of 30 competition-level problems. GPQA-Diamond is the high-quality subset of GPQA, comprising 198 expert-written multiple-choice questions across biology, physics, and chemistry. For AIME 2024, AIME 2025, and GPQA-Diamond, we evaluate on the full test sets to assess robustness under extremely challenging settings.

For SpecBench, since most tasks are open-ended, we adopt the LLM-judge protocol from FastChat~\cite{fastchat}, using GPT-4.1~\cite{gpt4} as the judge model to score responses from 0 to 10 on a per-turn basis. For the remaining benchmarks, we report accuracy.

% \input{Tables/table_new_benchmark}

% As shown in Table~\ref{tab:new_benchmark}, TriSpec consistently achieves stable speed‑up over the standard speculative decoding (SD) framework in both general‑domain and extremely difficult tasks. In terms of generation quality, on the SpecBench benchmark, the overall average score decreases by only 0.2 compared to the target model, substantially outperforming the proxy‑only setting. On HotpotQA, TriSpec exhibits almost no accuracy degradation. Even on Polymath, AIME24\&25 and GPQA-diamond, where the gap between the proxy and the target model is considerable, TriSpec maintains performance close to that of the target model, while delivering more than 30\% speed‑up over standard SD. Figure~\ref{fig:specbench} presents the score and speed‑up for each SpecBench category, further confirming TriSpec’s effectiveness across diverse task types.

% \begin{figure}
%     \centering
%     \includegraphics[width=\linewidth]{Pictures/radar.png}
%     \caption{Radar chart illustrating score and speed across all SpecBench categories. All experiments are conducted using models from the Qwen3 family.}
%     \label{fig:specbench}
% \end{figure}

\subsection{Experiments on Non-Reasoning Models}

We further evaluate TriSpec on non-reasoning models, using LLaMA-3.1-8B-Instruct as the target model and LLaMA-3.2-1B-Instruct as the proxy verifier. As shown in Table~\ref{tab:llama3_results}, TriSpec remains effective in this setting, maintaining accuracy close to target-only decoding while achieving faster verification than standard speculative decoding.

\begin{table*}[ht]
\centering
\captionsetup[table]{skip=5pt}
\setlength{\tabcolsep}{3pt} 
\caption{Accuracy and speedup results for original LLaMA3 family}
\label{tab:llama3_results}
\footnotesize
\begin{tabular}{lcccccccccc}
\toprule
& \multicolumn{2}{c}{GSM8K} & \multicolumn{2}{c}{MATH500} & \multicolumn{2}{c}{Gaokao2023en} & \multicolumn{2}{c}{Humaneval} & \multicolumn{2}{c}{MBPP}\\
\midrule
\multicolumn{1}{c}{Method} & Acc. & Speedup & Acc. & Speedup & Acc. & Speedup & Acc. & Speedup & Acc. & Speedup \\
\midrule
Single Model (Target, 8B) & 64\% & 1.00$\times$ & 34\% & 1.00$\times$ & 18\% & 1.00$\times$ & 66\% & 1.00$\times$ & 57\% & 1.00$\times$ \\
Single Model (Proxy, 1B) & 34\% & 2.54$\times$ & 26\% & 2.53$\times$ & 10\% & 2.50$\times$ & 46\% & 2.58$\times$ & 29\% & 2.53$\times$ \\
HASS & 64\% & 2.93$\times$ & 34\% & 3.27$\times$ & 18\% & 3.23$\times$ & 66\% & 3.52$\times$ & 57\% & 3.46$\times$ \\
HASS + TriSpec & 62\% & \textbf{3.17$\times$} & 34\% & \textbf{3.73$\times$} & 17\% & \textbf{3.66$\times$} & 66\% & \textbf{3.81$\times$} & 54\% & \textbf{3.76$\times$} \\
\bottomrule
\end{tabular}
\end{table*}

\subsection{Proxy selection in TriSpec}
Figure~\ref{fig:error_bars}(c) compares the token-level exact match ratio between Qwen3 models of varying parameter counts and the Qwen3-32B target model. Match ratios increase with model size, but the marginal gains diminish beyond the 1.7B scale. Since token-level alignment directly impacts a proxy's ability to assist the target in verification, this trend is a key factor in proxy selection. In particular, the 0.6B model surpasses the single-layer EAGLE3 drafter by only about 6\%, yielding relatively few local corrections. Conversely, the 4B model offers merely a 2.5\% increase over the 1.7B model, yet incurs substantially higher inference overhead. Balancing speed and alignment effectiveness, we adopt the 1.7B model as the proxy in our main experiments.

\section{Further Analysis of the Three Models in TriSpec}
\subsection{Standalone examples}
\label{sec:appendix-standalone}

\begin{figure}[ht]
    \centering
    \includegraphics[width=\linewidth]{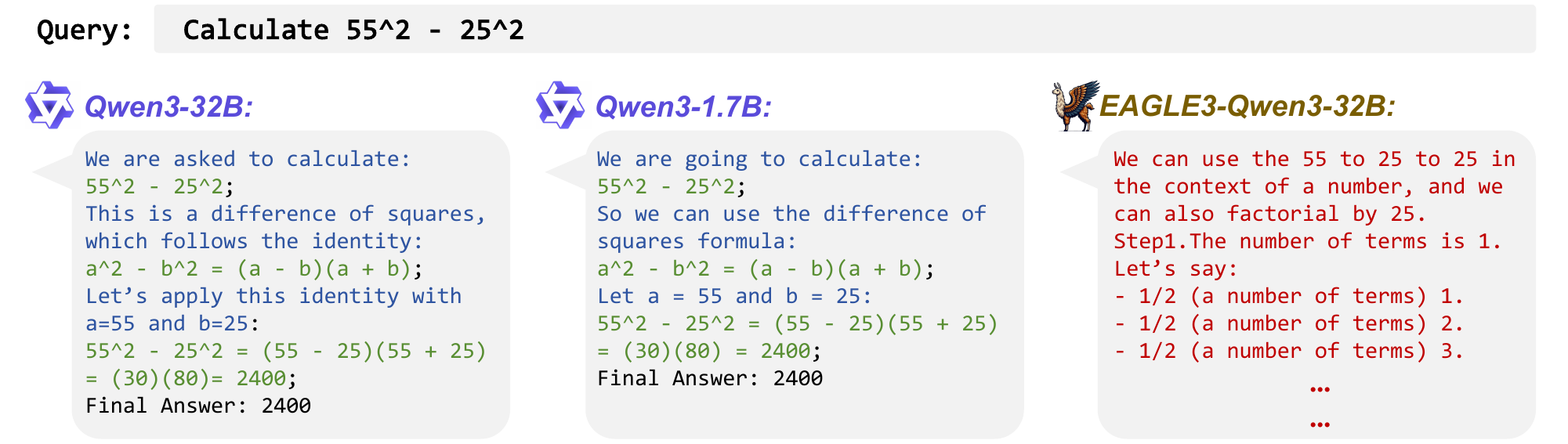}
    \caption{An example of standalone inference for the three models in TriSpec.}
    \label{fig:standalone_inference}
\end{figure}

Figure~\ref{fig:standalone_inference} presents a standalone inference example from the three models used in TriSpec within the Qwen3 family. The same-family smaller model (Qwen3-1.7B, proxy verifier) exhibits implicit alignment with the target model (Qwen3-32B) as reflected by their similar response format and language style. Meanwhile, the proxy also demonstrates reasonable standalone capability. Although its output contains token-level differences from the target, it still solves the problem correctly, suggesting that some proxy-target divergences are semantically tolerable and do not necessarily affect the final outcome.

In contrast, the single-layer drafter (EAGLE-3) is trained to mimic the target model but has much weaker contextual modeling capacity due to its substantially reduced attention computation. Without periodic target-model verification during generation, its standalone outputs can easily become illogical or inconsistent.

% {
% \captionsetup{labelfont={color=black}}
% }

\subsection{Token-level alignment experiments}
\label{sec:appendix-token-level-experiments}
\paragraph{Experiment details}
% To compare token-level alignment between the proxy verifier and the target model, we use the ShareGPT dataset as a reference. First, we use Qwen3-32B as the target model to generate autoregressive answers for each prompt. Then, we prefill the target completions using Qwen3-1.7B (the proxy) and record the proxy’s next-token prediction at each position. A token is considered an exact match if the proxy and target predictions coincide. For positions where they differ, we consult a stronger judge (Qwen3-235B) with top-k = 3 and top-p = 0.9 sampling parameters: if the proxy’s token is included in the set of tokens generated under these conditions, we deem it acceptable; otherwise, it is considered unacceptable. For both the proxy and the target model, temperature = 0 is used to ensure deterministic outputs.

To evaluate token-level alignment between the proxy verifier and the target model, we use ShareGPT as the reference dataset. We first generate autoregressive responses with Qwen3-32B as the target model. We then prefill the target completions with Qwen3-1.7B, the proxy verifier, and record the proxy's next-token prediction at each position. A token is counted as an exact match when the proxy and target predictions are identical. For non-matching positions, we further determine whether the proxy token can be categorized as a benign divergence. Specifically, we first check whether the proxy token lies within the target model's top-\(k\) candidates or top-\(p\) nucleus at the same position, with \(k=3\) and \(p=0.9\). If not, we consult Qwen3-235B as a stronger judge using the same top-\(k\) and top-\(p\) settings, mark the proxy token as benign if it appears in the resulting candidate set. All remaining mismatches are labeled as salient divergences. For both the proxy and target models, we use temperature \(=0\) to ensure deterministic outputs.

\begin{figure}[h]
    \centering
    \includegraphics[width=\linewidth]{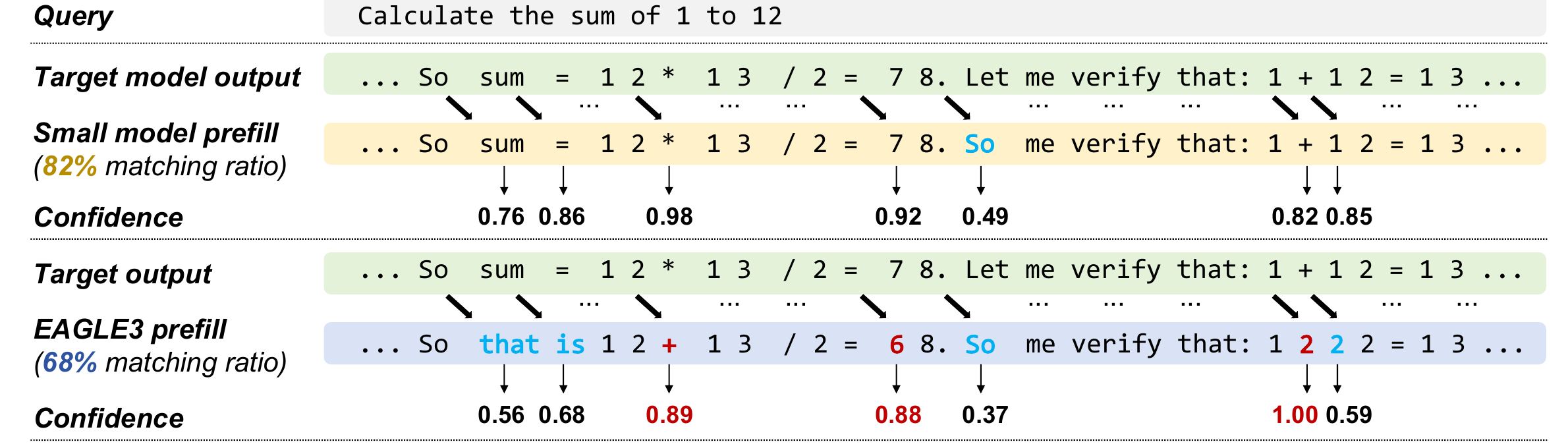}
    \caption{Token-level comparison of the three models in TriSpec. Blue highlights mark benign divergences, and red highlights mark salient divergences. The reported matching ratios are measured over all ShareGPT samples.}
    \label{fig:token_level_case}
\end{figure}

\paragraph{Case analysis.}
We further compare token-level alignment across the same-family small model and the single-layer EAGLE-3 drafter. Figure~\ref{fig:token_level_case} presents a representative case, showing token-level outputs from the three models together with their confidence scores. The proxy verifier typically assigns higher confidence to exact-match tokens and lower confidence to divergent tokens, indicating that its confidence provides an informative signal for prediction reliability. Overall, the proxy verifier shows stronger alignment with the target than the EAGLE-3 drafter, achieving an exact-match ratio of 82\% compared with 68\%. This advantage is particularly evident for decision-critical tokens, such as ``*'' and ``78'', where the proxy can locally correct errors made by the single-layer drafter. By contrast, due to its limited capacity, the single-layer drafter can be overconfident in erroneous predictions, making its confidence scores unreliable for deciding whether its outputs should be accepted.
% {
% \captionsetup{labelfont={color=black}}
% }

% \begin{figure}
%     \centering
%     \includegraphics[width=0.7\linewidth]{Pictures/rebuttal1.png}
%     \caption{\textcolor{blue}{Token-level exact match ratios of models with varying parameter scales and the EAGLE3 drafter, computed using Qwen3‑32B predictions as reference.}}
%     \label{fig:match_ratio}
% \end{figure}

\subsection{Attempts at relaxing verification}
\label{sec:appendix-relax-verification}

This work focuses on the verification side of speculative decoding and aims to accelerate generation by reducing the cost of strict target-model verification. Besides the proxy-based verification adopted in TriSpec, another natural direction is to relax verification constraints so as to increase the acceptance length. We examine two common relaxation strategies in the SD pipeline: (i) confidence filtering, which accepts a drafter token and skips target verification when the drafter confidence score, \(\max p_i\), exceeds a threshold; and (ii) top-$k$ verification, which accepts a drafter token if it appears among the target model's top-$k$ candidates. We evaluate these strategies using both single-layer drafters, i.e., HASS/EAGLE3 weights, and same-family smaller models as draft models. In these experiments, we use a chain-based draft structure with draft length set to 3.

\begin{figure}[h]
    \centering
    \includegraphics[width=0.9\linewidth]{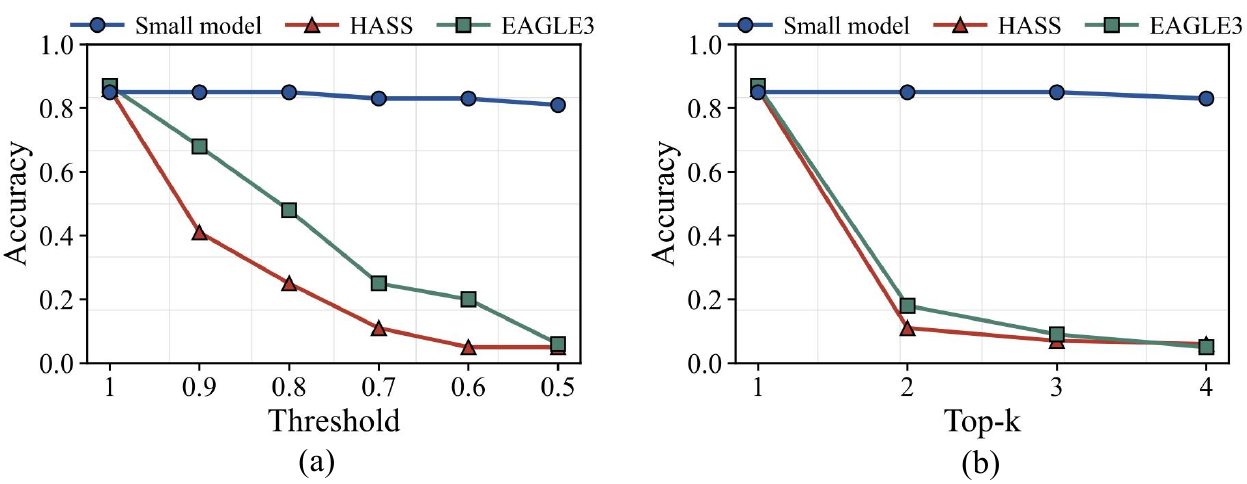}
    \caption{Verification relaxation in the Qwen3 family on MATH500. (a) Confidence filtering under different confidence thresholds. (b) Top-$k$ verification under different values of \(k\).}
    \label{fig:time_and_relaxation}
\end{figure}

\paragraph{Single-layer drafters are difficult to relax.}
As shown in Figure~\ref{fig:time_and_relaxation}(b), single-layer drafters suffer rapid accuracy degradation under both relaxation strategies. Confidence filtering may admit high-confidence but incorrect drafts, because capacity-limited single-layer drafters are often overconfident. Top-$k$ verification is also brittle, as it may accept near-optimal yet incorrect tokens that single-layer drafters frequently produce at decision-critical positions. As a result, relaxed verification is fragile for single-layer drafters: even mild relaxation can induce noticeable accuracy loss, limiting the attainable speedup. This trend is further confirmed by our benchmark results in Table~\ref{tab:model_performance}, where confidence filtering with a high threshold of 0.95 still struggles to preserve accuracy and often leads to repeated outputs, which we attribute to the single-layer drafter's unwarranted overconfidence discussed in Appendix~\ref{sec:appendix-token-level-experiments}. This motivates us to avoid drafter-side confidence signals when designing the routing strategy.

\paragraph{High-confidence divergences in same-family smaller models can be tolerated.}
For same-family smaller models, relaxing verification with an appropriate threshold does not cause significant accuracy loss. This end-to-end experiment shows that high-confidence divergences from smaller models are often tolerable, further supporting their use as proxy verifiers. A similar observation is also reported in Judge Decoding~\cite{judgedecoding}.

% {
% \captionsetup{labelfont={color=black}}
% }

\section{Detailed Error Analysis}
\label{app:error_analysis}

\subsection{First-order Task-level Performance-gap Decomposition}
\label{app:task_gap}

\paragraph{Task metric and common reference space.}
For an input $q\sim\mathcal D$, let $U_A(q,\xi_A)\in[0,1]$ denote method $A$'s score under a common normalized task metric, where $\xi_A$ contains its decoding randomness. Binary correctness for pass@1 is one example. We define $\Delta_A=\mathbb E[U_t]-\mathbb E[U_A]$, with expectations over inputs and the respective decoding randomness. This is a signed gap and can be negative.

We use a target-only trajectory as a common reference. A reference realization $\omega$ comprises the prompt, the finite target-only response, coupled random choices, and aligned auxiliary draft/proxy states supplying a potential proxy token at each reference history. The potential proxy token is a fixed reference candidate supplied by the proxy-side token decision rule on these aligned states, including speculative acceptance of a draft token or local correction, as applicable; it is not simply the proxy top-1 prediction. The reference histories, auxiliary states, random choices, and resulting candidates are fixed before varying $\lambda$, independently of whether proxy finalization occurs. Actual proxy-finalized outputs may differ across thresholds; identifying them with these fixed candidates is part of the first-order approximation, not an assertion that the routed output is invariant to $\lambda$. At each reference position, replace only the target-selected token with its potential proxy token, then continue target decoding using the same subsequent random choices and termination rule as in the unmodified reference. A position is task-critical precisely when this paired intervention changes the final task score. Once $\omega$ is fixed, this is a definite comparison, not an existential statement over alternative future randomness.

Let $\mathcal C(\omega)$ be the resulting set of task-critical positions and let $c=\mathbb E_\omega[|\mathcal C(\omega)|]$, assuming $0<c<\infty$. We use the pooled-position probability measure
\[
\Pr_{\mathrm{ref}}(A)
=\frac{\mathbb E_\omega\!\left[\sum_{i\in\mathcal C(\omega)}
\mathbf 1\{A(\omega,i)\}\right]}{c}.
\]
Thus, positions are pooled with their input probabilities; per-input event fractions are not averaged with equal weight. On this same space, $S$ denotes a salient divergence between the potential proxy token and the target token. The event $F_\lambda$ is the reference-model flag that the routing rule, evaluated on the aligned reference draft/proxy states, would finalize that decision through the proxy without target verification. All probabilities below are under $\Pr_{\mathrm{ref}}$.

Task-critical positions are latent analytical objects, not token classes identified during inference. Actual proxy-only and TriSpec histories can depart from the target reference, so mapping their decisions and routing events onto these aligned states is itself a first-order modeling approximation. In particular, $F_\lambda$ is not an exact statistic of actual TriSpec trajectories. The candidate-set criterion in Appendix~\ref{sec:appendix-token-level-experiments} makes $S$ a surrogate for potentially harmful decisions: it does not establish that every benign divergence is harmless or every salient divergence changes the task outcome.

\paragraph{First-order derivation.}
Define $p_\lambda=\Pr(F_\lambda)$ and $\mu_0=\Pr(S)$; when $p_\lambda>0$, define $\mu_\lambda=\Pr(S\mid F_\lambda)$. By the pooled-position definition, the expected reference counts of salient divergences and proxy-finalized salient divergences are exactly $c\mu_0$ and $c\Pr(F_\lambda\cap S)=cp_\lambda\mu_\lambda$, respectively, with the latter factorization used only when $p_\lambda>0$. These are counts in the reference model, not measured counts on the methods' diverging generation paths.

To relate these counts to task scores, we approximate the gap by the sum of signed contributions from salient divergences, neglecting contributions not captured by this surrogate, interactions among interventions, and changes to subsequent trajectories. We further assume a comparable mean signed contribution $\kappa$ for all reference salient divergences and the subset selected by $F_\lambda$. This comparability need not hold merely because routing uses a confidence margin. For $p_\lambda>0$ and $\mu_0>0$, the model gives
\[
\Delta_{\mathrm{proxy}}\approx c\mu_0\kappa,
\qquad
\kappa\approx\frac{\Delta_{\mathrm{proxy}}}{c\mu_0},
\qquad
\Delta_{\mathrm{tri}}\approx cp_\lambda\mu_\lambda\kappa
\approx p_\lambda\frac{\mu_\lambda}{\mu_0}\Delta_{\mathrm{proxy}},
\]
recovering Eq.~\eqref{eq:error_estimate}. Here $\kappa$ is a modeling coefficient for signed score differences, not a necessarily positive penalty. The derivation is an explanatory approximation, not a performance bound.

\paragraph{Probability interpretation and boundary cases.}
For $p_\lambda>0$ and $\mu_0>0$, the following identity is exact on the common reference space:
\begin{equation}
\label{eq:reference_rho}
\rho_\lambda
=p_\lambda\frac{\mu_\lambda}{\mu_0}
=\frac{\Pr(F_\lambda\cap S)}{\Pr(S)}
=\Pr(F_\lambda\mid S).
\end{equation}
Thus, $\rho_\lambda\in[0,1]$ is the conditional fraction of salient-divergence positions among the reference task-critical positions that remain proxy-finalized. It combines exposure and conditional divergence in one quantity; no additional filtering factor should be multiplied into it. The identity is exact, whereas $\Delta_{\mathrm{tri}}\approx\rho_\lambda\Delta_{\mathrm{proxy}}$ still depends on the first-order assumptions above. Only when $\Delta_{\mathrm{proxy}}>0$ and $\rho_\lambda<1$ does the model predict smaller task-level degradation. It does not guarantee that the actual signed gap contracts by this factor.

If $p_\lambda=0$ and $\mu_0>0$, define $\rho_\lambda=\Pr(F_\lambda\cap S)/\Pr(S)=0$ without defining $\Pr(S\mid F_\lambda)$; the reference proxy-finalized salient count is zero. Eq.~\eqref{eq:error_estimate} must not be evaluated through the undefined conditional probability in this case. If $\mu_0=0$, the ratio and $\Pr(F_\lambda\mid S)$ are undefined; if $c=0$, the pooled reference measure is unavailable. We do not apply the decomposition in these cases or infer that the actual performance gap vanishes. Finally, Figure~\ref{fig:latency_breakdown}(b) describes global token-level margin/divergence statistics and supplies qualitative evidence for filtering, not an estimate of $\rho_\lambda$ or the latent task-critical probabilities.

\subsection{Sequence-level Distributional Bound}
\label{app:sequence_bound}

\paragraph{Sequence distributions and token-level routing.}
Fix a common prompt, a sampling configuration (including temperature and any top-$k$/top-$p$ truncation), and a horizon $T$. We retain the first $T$ generated tokens, padding terminated responses with an absorbing EOS symbol. Let $P_\lambda^{(T)}$ and $P_t^{(T)}$ be the resulting TriSpec and target-only distributions, and $H_i=Y_{<i}$ the generated history before position $i$. All distributions in this subsection are conditional on the fixed prompt. Unlike Appendix~\ref{app:task_gap}, this analysis uses the actual generation law, not the target-reference approximation. KL uses natural logarithms, and $\mathrm{TV}(P,Q)=\tfrac12\sum_y|P(y)-Q(y)|$.

Under TriSpec, let $Z_i=1$ if token $i$ is finalized through the proxy and $Z_i=0$ if it is finalized through the target. This distinction concerns the token's final decision source, not whether its decoding round invokes the target. In particular, proxy-finalized prefix tokens pruned before target escalation have $Z_i=1$. At each output history $h$ with positive probability under TriSpec, define
\[
\begin{aligned}
\alpha_{\lambda,i}(h)&=\Pr_\lambda(Z_i=1\mid H_i=h),\\
r_{\lambda,i}(\cdot\mid h)&=\Pr_\lambda(Y_i\in\cdot\mid H_i=h,Z_i=1),\\
s_{\lambda,i}(\cdot\mid h)&=\Pr_\lambda(Y_i\in\cdot\mid H_i=h,Z_i=0),
\end{aligned}
\]
and let $p_i^{(t)}(\cdot\mid h)$ be the target sampling kernel under the same configuration, evaluated at that history. This kernel is specified by the target decoding policy even at histories unreachable under target-only generation; on histories with positive target probability, it is the target-only conditional. Internal drafts, cache states, and algorithm randomness are marginalized conditional on the displayed events. Consequently, $r_{\lambda,i}$ includes acceptance, correction, and routing selection, and is not generally the proxy's raw softmax distribution. For deterministic EOS padding, set $Z_i=0$ and use the EOS point mass for the output conditional. Conditionals on zero-probability routes may be assigned arbitrarily, and their weighted KL contributions are defined as zero.

Writing $q_{\lambda,i}$ for the actual TriSpec next-token conditional, the law of total probability gives
\begin{equation}
\label{eq:routing_mixture}
q_{\lambda,i}(\cdot\mid h)
=(1-\alpha_{\lambda,i}(h))s_{\lambda,i}(\cdot\mid h)
+\alpha_{\lambda,i}(h)r_{\lambda,i}(\cdot\mid h).
\end{equation}
Routing depends on draft/proxy outcomes, so conditioning on the target route can alter its output law. Standard speculative verification alone therefore does not establish $s_{\lambda,i}=p_i^{(t)}$ after this selection.

\paragraph{KL decomposition and TV bound.}
Expectations are under TriSpec's joint generation law; history-only expectations are written $\mathbb E_{P_\lambda}$. Conditional distributions inside divergences are evaluated at $H_i$. Let $N_F(\lambda)=\sum_{i=1}^T Z_i$ count proxy-finalized tokens, excluding EOS padding. Then
\[
\mathbb E[N_F(\lambda)]
=\mathbb E_{P_\lambda}\!\left[\sum_{i=1}^T\alpha_{\lambda,i}(H_i)\right].
\]
Define the accumulated target-route contribution
\[
B_\lambda=
\mathbb E_{P_\lambda}\!\left[\sum_{i=1}^T
(1-\alpha_{\lambda,i}(H_i))\,
\mathrm{KL}\!\left(s_{\lambda,i}(\cdot\mid H_i)\middle\|p_i^{(t)}(\cdot\mid H_i)\right)\right].
\]
For $\mathbb E[N_F(\lambda)]>0$, define the exposure-weighted proxy-route divergence
\[
\varepsilon_\lambda=
\frac{\displaystyle\mathbb E_{P_\lambda}\!\left[\sum_{i=1}^T
\alpha_{\lambda,i}(H_i)\,
\mathrm{KL}\!\left(r_{\lambda,i}(\cdot\mid H_i)\middle\|p_i^{(t)}(\cdot\mid H_i)\right)\right]}
{\mathbb E[N_F(\lambda)]}.
\]
If $\mathbb E[N_F(\lambda)]=0$, every proxy-route weight is zero almost surely; its accumulated weighted KL contribution is defined as zero. We use $\mathbb E[N_F(\lambda)]\varepsilon_\lambda=0$ as shorthand for that contribution, without dividing by zero.

The autoregressive KL chain rule and convexity in the first argument yield
\begin{equation}
\label{eq:sequence_kl_derivation}
\begin{aligned}
\mathrm{KL}\!\left(P_\lambda^{(T)}\middle\|P_t^{(T)}\right)
&=\mathbb E_{P_\lambda}\!\left[\sum_{i=1}^T
\mathrm{KL}\!\left(q_{\lambda,i}\middle\|p_i^{(t)}\right)\right]\\
&\leq\mathbb E_{P_\lambda}\!\left[\sum_{i=1}^T
\left\{(1-\alpha_{\lambda,i})\mathrm{KL}\!\left(s_{\lambda,i}\middle\|p_i^{(t)}\right)
+\alpha_{\lambda,i}\mathrm{KL}\!\left(r_{\lambda,i}\middle\|p_i^{(t)}\right)\right\}\right]\\
&=B_\lambda+\mathbb E[N_F(\lambda)]\varepsilon_\lambda.
\end{aligned}
\end{equation}
Pinsker's inequality, $\mathrm{TV}(P,Q)\leq\sqrt{\mathrm{KL}(P\|Q)/2}$, gives the second inequality in Eq.~\eqref{eq:sequence_deviation}. For a finite-valued bound, require $P_\lambda^{(T)}\ll P_t^{(T)}$ and finite weighted KL terms. Absolute continuity here is equivalent to requiring that, at every TriSpec-positive history, the support of $q_{\lambda,i}$ (and hence of each positive-weight route conditional) must be contained in that of $p_i^{(t)}$. This also makes all such histories reachable under the target policy. If this support condition fails, sequence KL is infinite; the chain rule and inequalities are understood in the extended-real sense and supply no finite KL guarantee. Positive temperature does not by itself resolve this issue when top-$k$/top-$p$ truncation changes the support.

\paragraph{Simplified case and interpretation.}
The general bound does not assume target-route conditional consistency. If, additionally, $s_{\lambda,i}(\cdot\mid H_i)=p_i^{(t)}(\cdot\mid H_i)$ almost surely on active target routes, then $B_\lambda=0$ and
\[
\mathrm{KL}\!\left(P_\lambda^{(T)}\middle\|P_t^{(T)}\right)
\leq\mathbb E[N_F(\lambda)]\varepsilon_\lambda,
\qquad
\mathrm{TV}\!\left(P_\lambda^{(T)},P_t^{(T)}\right)
\leq\sqrt{\frac{\mathbb E[N_F(\lambda)]\varepsilon_\lambda}{2}}.
\]
When $\mathbb E[N_F(\lambda)]=0$, the general bound still contains $B_\lambda$. Equality with the target distribution follows only with further conditions such as target-route consistency, or when disabling proxy finalization restores standard lossless SD under the same sampling configuration.

$B_\lambda$ measures target-route conditional divergence, including routing-induced selection effects; it is neither computational overhead nor the actual output error. Convexity can make the bound loose: individual route divergences may be positive even when their mixture equals the target conditional. Since $B_\lambda$ and $\varepsilon_\lambda$ are not directly estimated here, the result characterizes sources of bias rather than establishing that actual sequence-level bias is small. Increasing $\lambda$ tightens the margin eligibility test at a fixed state, but changes exposure, conditional distributions, and visited histories; no monotonic decrease in sequence KL follows without additional conditions. Margin confidence therefore supplies no distribution-free guarantee.

Under deterministic greedy decoding with a fixed prompt, different $T$-token outputs give two distinct point masses: sequence-level TV is one and KL is infinite. Token-level disagreement can provide empirical reliability evidence, but low local disagreement does not by itself establish small sequence TV. Likewise, margin-bin disagreement and target-support statistics are not direct estimates of the conditional KL terms above. These limitations do not alter the bound, but restrict its informativeness.

\section{Case Study of Complete Generation}
\begin{figure}[ht]
    \centering
    \includegraphics[width=0.95\linewidth]{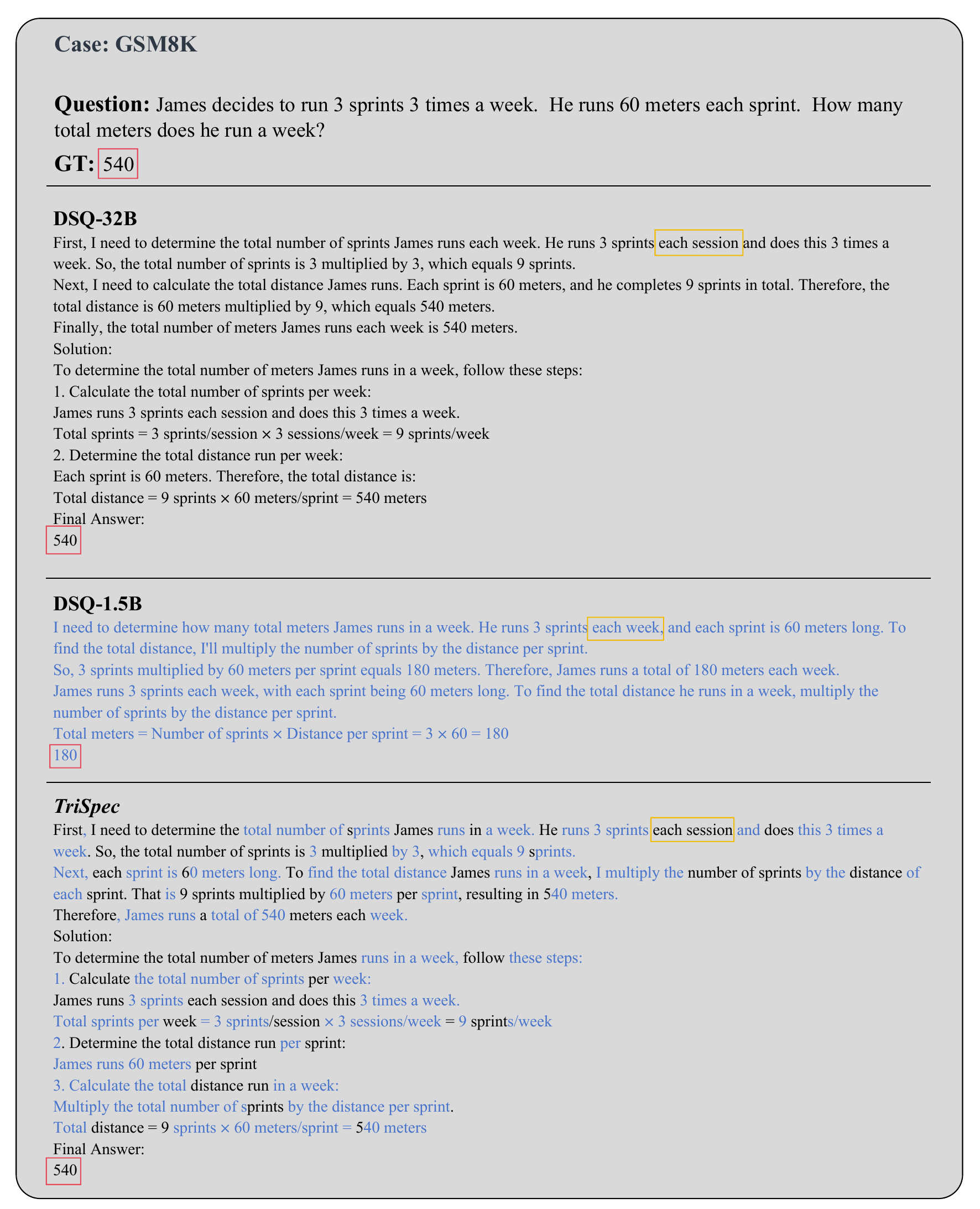}
    \caption{Case study of TriSpec in GSM8K.}
    \label{fig:case_study}
\end{figure}

We present a TriSpec case study of complete generation and compare it with target-only and proxy-only decoding results(Figure~\ref{fig:case_study}). Tokens in blue denote outputs verified or generated by the proxy model (DSQ-1.5B), while tokens in black indicate those verified or generated by the target model (DSQ-32B). In this example, over half of the tokens are validated by the proxy model, highlighting TriSpec’s efficiency. When the proxy alone would err, TriSpec attains correctness by involving the target, preserving accuracy.

Specifically, the yellow boxes highlight disagreements between the two models. The proxy incorrectly predicts “He runs 3 sprints each week” instead of the correct “He runs 3 sprints each session.” In TriSpec inference, however, the proxy expresses uncertainty at this point and defers to the target model, thereby avoiding the error and showcasing strong robustness.

\section{Limitations}
\label{sec:limitation}
TriSpec relaxes the exact losslessness guarantee of standard speculative decoding by allowing the proxy model to finalize part of the verification process. Although our theoretical analysis and extensive experiments suggest that the resulting accuracy degradation is limited, such approximate verification cannot guarantee identical behavior on novel tasks and model families. In addition, TriSpec requires the proxy and target models to share a tokenizer for direct token-level verification, which limits proxy selection, particularly across different model families. Finally, introducing an additional proxy model brings new system-level challenges, making optimized deployment in batched serving and production inference systems an important direction for future work.

\section{Licenses for existing assets}
\label{sec:license}
Our paper builds on several existing models, datasets, code repositories, and external services. We acknowledge the original creators of these assets and summarize their licenses below:

\textbf{Qwen3-32B} is available at \url{https://huggingface.co/Qwen/Qwen3-32B} under Apache License 2.0.

\textbf{Qwen3-1.7B} is available at \url{https://huggingface.co/Qwen/Qwen3-1.7B} under Apache License 2.0.

\textbf{Qwen3-235B} is available at \url{https://huggingface.co/Qwen/Qwen3-235B-A22B} under Apache License 2.0.

\textbf{DeepSeek-R1-Distill-Qwen-32B} is available at \url{https://huggingface.co/deepseek-ai/DeepSeek-R1-Distill-Qwen-32B} under MIT License.

\textbf{DeepSeek-R1-Distill-Qwen-1.5B} is available at \url{https://huggingface.co/deepseek-ai/DeepSeek-R1-Distill-Qwen-1.5B} under MIT License.

\textbf{Llama-3.1-8B-Instruct} is available at \url{https://huggingface.co/meta-llama/Llama-3.1-8B-Instruct} under Llama 3.1 License.

\textbf{Llama-3.2-1B-Instruct} is available at \url{https://huggingface.co/meta-llama/Llama-3.2-1B-Instruct} under Llama 3.2 License.

\textbf{DeepSeek-R1-Distill-Llama-70B} is available at \url{https://huggingface.co/deepseek-ai/DeepSeek-R1-Distill-Llama-70B} under MIT License.

\textbf{DeepSeek-R1-Distill-Llama-8B} is available at \url{https://huggingface.co/deepseek-ai/DeepSeek-R1-Distill-Llama-8B} under MIT License.

\textbf{ShareGPT} is available at \url{https://huggingface.co/datasets/anon8231489123/ShareGPT_Vicuna_unfiltered} under Apache License 2.0.

\textbf{UltraChat-200K} is available at \url{https://huggingface.co/datasets/HuggingFaceH4/ultrachat_200k} under MIT License.

\textbf{GSM8K} is available at \url{https://huggingface.co/datasets/openai/gsm8k} under MIT License.

\textbf{MATH500} is available at \url{https://huggingface.co/datasets/HuggingFaceH4/MATH-500} under MIT License.

\textbf{Gaokao-2023-EN} is available at \url{https://github.com/OpenLMLab/GAOKAO-Bench} under Apache License 2.0.

\textbf{HumanEval} is available at \url{https://huggingface.co/datasets/openai/openai_humaneval} under MIT License.

\textbf{MBPP} is available at \url{https://huggingface.co/datasets/google-research-datasets/mbpp} under CC-BY-4.0 License.

\textbf{SpecBench} is available at \url{https://github.com/hemingkx/Spec-Bench} under Apache License 2.0.

\textbf{HotpotQA} is available at \url{https://huggingface.co/datasets/hotpotqa/hotpot_qa} under CC-BY-SA-4.0 License.

\textbf{PolyMath} is available at \url{https://huggingface.co/datasets/Qwen/PolyMath} under Apache License 2.0.

\textbf{AIME 2024} is available at \url{https://huggingface.co/datasets/math-ai/aime24} under Apache License 2.0.

\textbf{AIME 2025} is available at \url{https://huggingface.co/datasets/math-ai/aime25} under Apache License 2.0.

\textbf{GPQA-Diamond} is available at \url{https://github.com/idavidrein/gpqa} under MIT License.

\textbf{EAGLE} is available at \url{https://github.com/SafeAILab/EAGLE} under Apache License 2.0.

\textbf{HASS} is available at \url{https://github.com/HArmonizedSS/HASS} under Apache License 2.0.

\textbf{FastChat} is available at \url{https://github.com/lm-sys/FastChat} under Apache License 2.0.

\end{document}